\documentclass[runningheads]{llncs}

\usepackage{graphicx}
\usepackage{amsmath,amssymb} 
\usepackage{color}
\usepackage{subfigure}
\usepackage{booktabs}
\usepackage{makecell}
\usepackage{multirow}
\usepackage{bm}
\usepackage{marvosym}
\usepackage{url}

\begin{document}
\pagestyle{headings}
\mainmatter

\def\ACCV20SubNumber{825}  

\title{Lightweight Single-Image Super-Resolution Network with Attentive Auxiliary Feature Learning} 
\titlerunning{A$^2$F: An Efficient SR Network}
%
\author{Xuehui Wang\inst{1} \and
Qing Wang\inst{1} \and
Yuzhi Zhao\inst{2} \and
Junchi Yan\inst{3} \and
Lei Fan\inst{4}  \and \\
Long Chen\inst{1}\textsuperscript{(\Letter)} }
\authorrunning{X. Wang et al.}
%
\institute{
School of Data and Computer Science, Sun Yat-sen University, Guangzhou, China \\
\email{wangxh228@mail2.sysu.edu.cn, chenl46@mail.sysu.edu.cn} \and
City University of Hong Kong, Hong Kong, China \and
Shanghai Jiao Tong University, Shanghai, China\and
Northwestern University, Evanston, USA }

\maketitle

\begin{abstract}
Despite convolutional network-based methods have boosted the performance of single image super-resolution (SISR), the huge computation costs restrict their practical applicability. In this paper, we develop a computation efficient yet accurate network based on the proposed attentive auxiliary features (A$^2$F) for SISR. Firstly, to explore the features from the bottom layers, the auxiliary feature from all the previous layers are projected into a common space. Then, to better utilize these projected auxiliary features and filter the redundant information, the channel attention is employed to select the most important common feature based on current layer feature. We incorporate these two modules into a block and implement it with a lightweight network. Experimental results on large-scale dataset demonstrate the effectiveness of the proposed model against the state-of-the-art (SOTA) SR methods. Notably, when parameters are less than 320k, A$^2$F outperforms SOTA methods for all scales, which proves its ability to better utilize the auxiliary features. Codes are available at \url{https://github.com/wxxxxxxh/A2F-SR}.
\end{abstract}

\section{Introduction}
Convolutional neural network (CNN) has been widely used for single image super-resolution (SISR) since the debut of SRCNN~\cite{SRCNN}. Most of the CNN-based SISR models~\cite{VDSR,EDSR,SRDensenet,WDSR,WangmenZuo,RCAN} are deep and large. However, in the real world, the models often need to be run efficiently in embedded system like mobile phone with limited computational resources~\cite{chen,videoenhancement1,medical1,medical2,tracking2,monitoring}. Thus, those methods are not proper for many practical SISR applications, and lightweight networks have been becoming an important way for practical SISR. Also, the model compression techniques can be used in lightweight architecture to further reduce the parameters and computation. However, before using model compression techniques (e.g. model pruning), it is time-consuming to train a large model and it also occupies more memory. This is unrealistic for some low budget devices, so CNN-based lightweight SISR methods become increasingly popular because it can be regarded as an image preprocessing or postprocessing instrument for other tasks~\cite{CSL,surrounding,shen2019deep,scrdet,learning}.
\begin{figure}[tb!]
   \centering
   \includegraphics[width=0.6\textwidth]{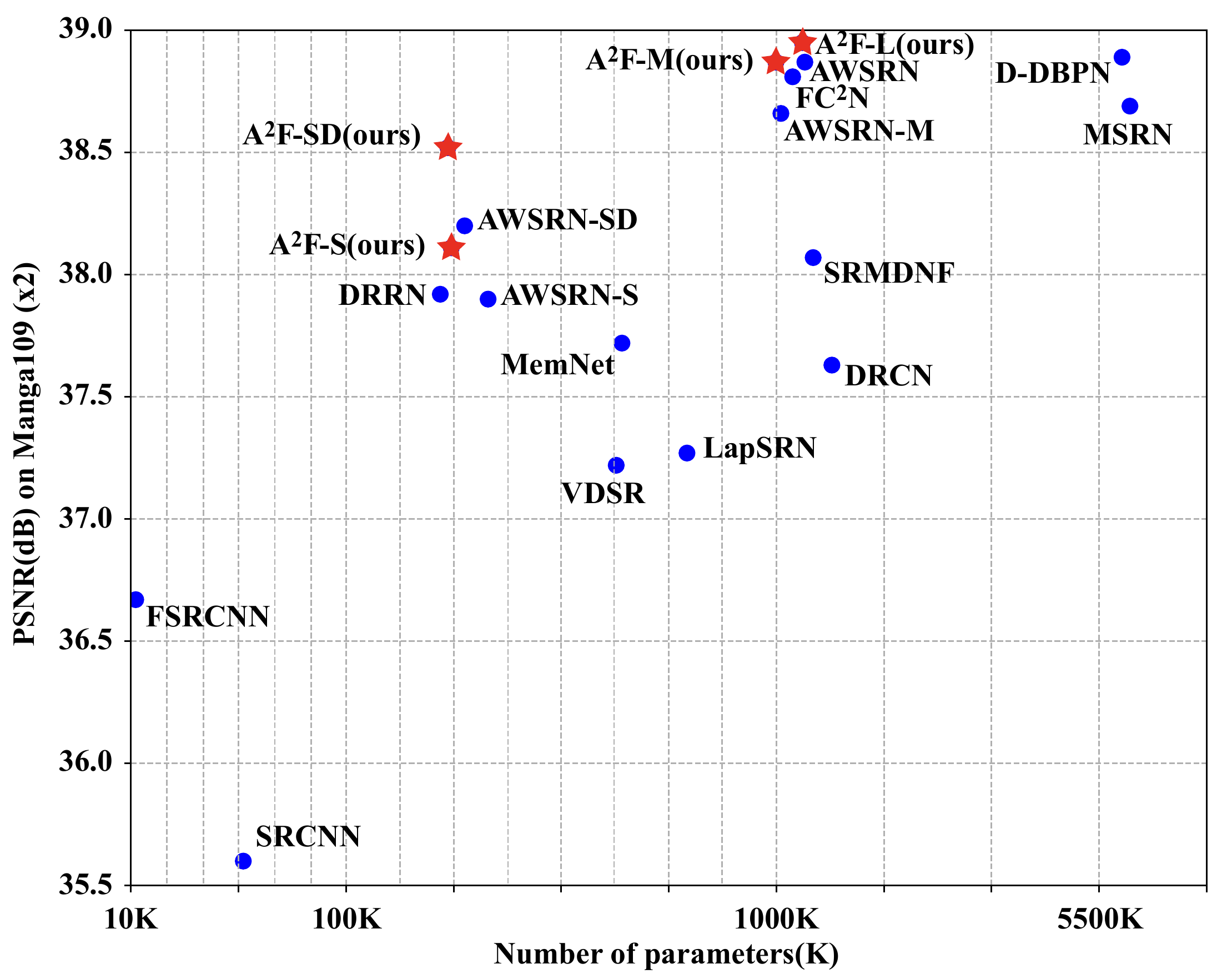} 
   \caption{Cost-effectiveness comparison between the proposed A$^2$F model variants (A$^2$F-S, A$^2$F-SD, A$^2$F-M, A$^2$F-L) with other methods on the Manga109~\cite{Manga109} on $\times$2 scale. The proposed models can achieve high PSNR with fewer parameters. Note that MSRN~\cite{MSRN} and D-DBPN~\cite{DBPN} are large models.}
   \label{effComp}

\end{figure}

One typical strategy is to reduce the parameters~\cite{FSRCNN,DRCN,ESPCN,DRRN}. Moreover, the network architecture is essential for lightweight SISR models. Generally, methods of designing architectures can be categorized into two groups. One is based on neural architecture search. MoreMNA-S and FALSR~\cite{FALSR,MoreMNAS} adopt the evolutionary algorithm to search efficient model architectures for lightweight SISR. The other is to design the models manually~\cite{AWSRN,FC2N}. These methods all utilize features of previous layers to better learn the features of the current layer, which reflect that auxiliary features can boost the performance of lightweight models. However, these methods do not fully use all the features of previous layers, which possibly limits the performance.

\begin{figure*}[tb!]
  \centering
  \includegraphics[width=0.95\textwidth]{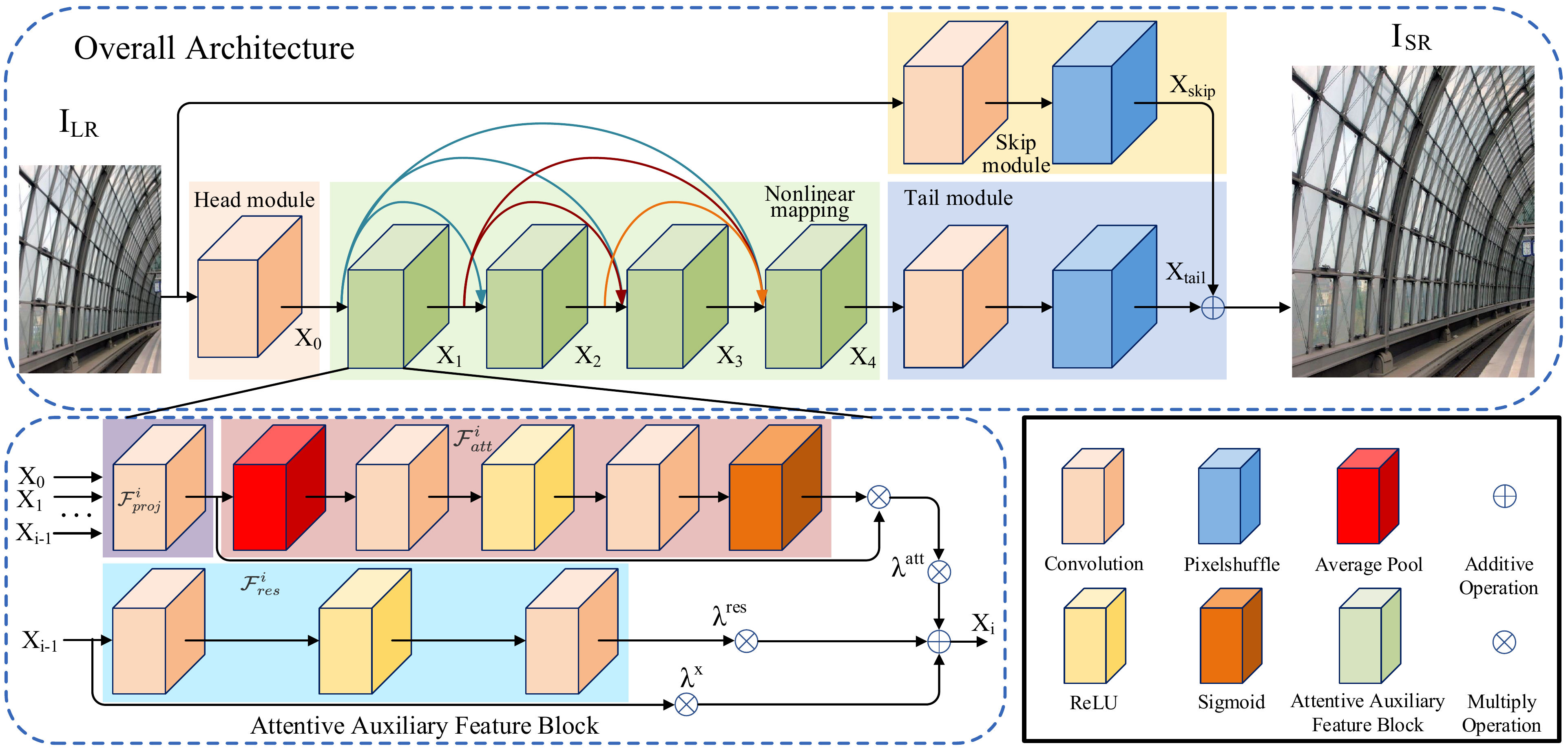} 
  \caption{The architecture of A$^2$F with 4 attentive auxiliary feature blocks. The architecture of A$^2$F with more attentive auxiliary feature blocks is similar. Note that 1$\times$1 convolution kernel is used to project the auxiliary features and learn the importance of different channels of projected features. The convolution kernels elsewhere are all $3 \times 3$. The input is the LR image and the output is the predicted HR image. Pixelshuffle~\cite{pixelshuffle} is used to upsample the features to the high-resolution image with target size.}
  \label{netArch}
  
\end{figure*}

Directly combining the auxiliary features with current features is conceptually problematic as features of different layers are often embedded in different space. Thus, we use the projection unit to project the auxiliary features to a common space that is suitable for fusing features. After projected to a common space, these projected features may not be all useful for learning features of the current layer. So we adopt the channel attention to make the model automatically assign the importance to different channels. The projection unit and channel attention constitute the proposed attentive auxiliary feature block. We term our model that consists of \textbf{A}ttentive \textbf{A}uxiliary \textbf{F}eature blocks as A$^2$F since it utilizes the auxilary features and the attention mechanism. Figure~\ref{effComp} gives the comparison between different models on Manga109~\cite{Manga109} dataset with a upscale factor of 2. As shown in Figure~\ref{effComp}, models of our A$^2$F family can achieve better efficiency than current SOTA methods~\cite{AWSRN,FC2N}. Figure~\ref{netArch} describes the architecture of A$^2$F with four attentive auxiliary feature blocks. Our main contributions are given below:

\begin{itemize}
\item We handle the super resolution task from a new direction, which means we discuss the benefit brought by auxiliary features in the view of how to recover multi-frequency through different layers. Thus, we propose the attentive auxiliary feature block to utilize auxiliary features of previous layers for facilitating features learning of the current layer. The mainstay we use the channel attention is the dense auxiliary features rather than the backbone features or the sparse skip connections, which is different from other works.
\item Compared with other lightweight methods especially when the parameters are less than 1000K, we outperform all of them both in PSNR and SSIM but have fewer parameters, which is an enormous trade-off between performance and parameters. In general, A$^2$F is able to achieve better efficiency than current state-of-the-art methods~\cite{FC2N,AWSRN,SRFBN}.
\item Finally, we conduct a thorough ablation study to show the effectiveness of each component in the proposed attentive auxiliary feature block. We release our PyTorch implementation of the proposed method and its pretrained models together with the publication of the paper.
\end{itemize}

\section{Related Work}
Instead of powerful computers with GPU, embedded devices usually need to run a super resolution model. As a result, lightweight SR architectures are needed and have been recently proposed.

One pioneering work is SRCNN~\cite{SRCNN} which contains three convolution layers to directly map the low-resolution (LR) images to high-resolution (HR) images. Subsequently, a high-efficiency SR model named ESPCNN~\cite{ESPCN} was introduced, which extracts feature maps in LR space and contains a sub-pixel convolution layer that replaces the handcrafted bicubic filter to upscale the final LR map into the HR images. DRRN~\cite{DRRN} also had been proposed to alleviate parameters by adopting recursive learning while increasing the depth. Then CARN~\cite{CARN} was proposed to obtain an accurate but lightweight result. It addresses the issue about heavy computation by utilizing the cascading mechanism for residual networks. More recently, AWSRN~\cite{AWSRN} was designed to decrease the heavy computation. It applies the local fusion block for residual learning. For lightweight network, it can remove redundancy scale branches according to the adaptive weights.

Feature fusion has undergone its tremendous progress since the ResNet~\cite{Resnet} was proposed, which implies the auxiliary feature is becoming the crucial aspect for learning. The full utilization of the auxiliary feature was adopted in DenseNet~\cite{densenet}. The authors take the feature map of each former layer into a layer, and this alleviates the vanishing gradient problem. SR methods also make use of auxiliary features to improve performance, such as~\cite{VDSR,DRRN,MemNet,RCAN,RDN}. The local fusion block of AWSRN~\cite{AWSRN} consists of concatenated AWRUs and a LRFU. Each output of AWRUs is combined one by one, which means a dense connection for a block. A novel SR method called FC$^2$N was presented in~\cite{FC2N}. A module named GFF was devised through implementing all skip connections by weighted channel concatenation, and it also can be considered as the auxiliary feature.

As an important technique for vision tasks, attention mechanism~\cite{attention} can automatically determine which component is important for learning. Channel attention is a type of attention mechanism, which concentrates on the impact of each feature channel. SENet~\cite{SENet} is a channel attention based model in the image classification task. In the domain of SR, RCAN~\cite{RCAN} had been introduced to elevate SR results by taking advantage of interdependencies among channels. It can adaptively rescale features according to the training target.

In our paper, auxiliaty features are not fully-dense connections, which indicates it is not dense in one block. We expect that each block can only learn to recover specific frequency information and provide auxiliary information to the next block. There are two main differences compred with FC$^2$N and AWSRN. One is that for a block of A$^2$F, we use the features of ALL previous blocks as auxiliary features of the current block, while FC$^2$N and AWSRN use the features of a FIXED number of previous blocks. The second is that we adopt channel attention to decide how to transmit different informations to the next block, but the other two works do not adopt this mechanism.

\section{Proposed Model}
\subsection{Motivation and  Overview}
Our method is motivated by an interesting fact that many CNN based methods~\cite{CARN,EDSR,FC2N} can reconstruct the high frequency details from the low resolution images hierachically, which indicates that different layers learn the capacity of recovering multi-frequency information. However, stacking more layers increases the computation burden and higher frequency information is difficult to regain. So we aim to provide a fast, low-parameters and accurate method that can restore more high frequency details on the basis of ensuring the accuracy of low frequency information reconstruction. According to this goal, we have the following observations:

\begin{itemize}
\item To build a lightweight network, how to diminish parameters and the multiply operation is essential. Generally, we consider reducing the depth or the width of the network, performing upsampling operation at the end of the network and adopting small kernel to reach this target. It also brings a new issue that a shallow network (i.e. fewer layers and fewer channels in each layer) can not have an excellent training result due to the lower complexity of the model, which also can be considered as an under-fitting problem.

\item For the limited depth and width of the network, feature reusing is the best way to solve the issue. By this way, the low-frequency information can be transmitted to the next layer easily and it is more useful to combine multi-level low-frequency features to obtain accurate high-frequency features. Thus, more features benefitting to recover high-frequency signal will circulate across the entire network. It will promote the capacity of learning the mapping function if the network is shallow.

\item We also consider another problem that the impact of multi-frequency information should be different when used for the learning of high frequency features. As the depth of the layer becomes deeper, effective information of the last layer provided for current layer is becoming rarer, because the learning of high frequency features is more and more difficult. So how to combine the information of all the previous layers to bring an efficient result is important and it should be dicided by the network.
\end{itemize}

Based on these observations, we design the model by reusing all features of the preceding layers and then concating them directly along channels like~\cite{densenet} in a block. Meanwhile, to reduce the disturbance brought by the redundant information when concating all of channels and adaptively obtain the multi-frequency reconstruction capability of different layers, we adopt the same-space attention mechanism in our model, which can avoid the situation that features from different space would cause extraodinary imbalance when computing the attention weight.

\begin{table}[!htbp]
  \centering
  \caption{Configurations of the proposed method. We set $stride=1$ for every convolutional operation to keep the same size in each layer. $i$ indicates the $i$-th A$^2$F module and $p$ means the scale factor. For the A$^2$F-SD model, we change the channels that are 32 in other models to 16 for each $\mathcal{F}$.}
  \smallskip
  \resizebox{0.58\textwidth}{!}{
    \begin{tabular}{|c|c|c|c|}
      \hline
      \multicolumn{1}{|c|}{   Function   } & Details & Kernel & \makecell[c]{Channels\\(Input, Output)}\\
      \hline
      $\mathcal{F}_{head}$ & Convolution& $3\times3$& (3, 32) \\
      \hline
      \multirow{2}{*}{$\mathcal{F}_{skip}$} & Convolution & $3\times3$ & (3, $p*p*3$) \\
      & PixelShuffle & - & - \\
      \hline
      $\mathcal{F}^i_{proj}$ & Convolution & $1\times1$ & ($i*32$, 32) \\
      \hline
      \multirow{5}{*}{$\mathcal{F}^i_{att}$} & Adaptive AvgPool & - & - \\
      & Convolution & $1\times1$ &(32, 32) \\
      & ReLU & - & - \\
      & Convolution & $1\times1$ &(32, 32) \\
      & Sigmoid & - & - \\
      \hline
      \multirow{3}{*}{$\mathcal{F}^i_{res}$} & Convolution & $3\times3$ & (32, 128) \\
      & ReLU & - & - \\
      & Convolution & $3\times3$ & (128, 32) \\
      \hline
      \multirow{2}{*}{$\mathcal{F}_{tail}$} & Convolution & $3\times3$ & (32, $p*p*3$) \\
      & PixelShuffle & - & - \\
      \hline
      
  \end{tabular}}
  
  \label{networkTable}
  
\end{table}

\subsection{Overall Architecture}
As shown in Figure~\ref{netArch}, the whole model architecture is divided into four components: head module, nonlinear mapping, skip module and tail module. Detailed configuration of each component can be seen in Table~\ref{networkTable}. We denote the low resolution and the predicted image as $I_{LR}$ and $I_{SR}$, respectively. The input is first processed by the head module $\mathcal{F}_{head}$  to get the features $x_0$:
\begin{equation}
x_0 = \mathcal{F}_{head}(I_{LR}),
\end{equation}
and $\mathcal{F}_{head}$ is just one 3$\times$3 convolutional layer (Conv). We do not use $1\times1$ Conv in the first layer for it can not capture the spatial correlation and cause a information loss of the basic low frequency. The reason why we use a $3\times3$ Conv rather than a $5\times5$ Conv is twofold: a) $3\times3$ Conv can use fewer parameters to contribute to the lightweight of the network. b) It is not suitable to employ kernels with large receptive field in the task of super-resolution, especially for the first layer. Recall that each pixel in downsampled image corresponds to a mini-region in the original image. So during the training, large receptive field may introduce irrelevant information.

Then the nonlinear mapping which consists of $L$ stacked attentive auxiliary feature blocks is used to further extract information from $x_0$. In the $i_{th}$ attentive auxiliary feature block, the features $x_i$ is extracted from all the features of the previous blocks $x_0, x_1, x_2,...,x_{i-1}$:
\begin{equation}
   x_{i} = g^{i}_{AAF}(x_0,x_1,...,x_{i-1}),
\end{equation}
where $g^{i}_{AAF}$ denotes attentive auxiliary feature block $i$.

After getting the features $x_{L}$ from  the last attentive auxiliary feature block, $\mathcal{F}_{tail}$, which is a 3$\times$3 convolution layer followed by a pixelshuffle layer~\cite{ESPCN}, is used to upsample $x_{L}$ to the features $x_{tail}$ with targe size:
\begin{equation}
   x_{tail} = \mathcal{F}_{tail}(x_L).
\end{equation}
We design this module to integrate the multi-frequency information produced by different blocks. It also correlates channels and spatial correlation, which is useful for pixelshuffle layer to rescale the image.

To make the mapping learning easier and introduce the original low frequency information to keep the accuracy of low frequency, the skip module $\mathcal{F}_{skip}$, which has the same component with $\mathcal{F}_{tail}$, is adopted to get the global residual information $x_{skip}$:
\begin{equation}
   x_{skip} = \mathcal{F}_{skip}(I_{LR}).
\end{equation}
Finally, the target $I_{SR}$ is obtained by adding $x_{skip}$ and $x_{tail}$:
\begin{equation}
   I_{SR} = x_{tail} \oplus x_{skip}.
\end{equation}
where $\oplus$ denotes the element-wise add operation.

\subsection{Attentive Auxiliary Feature Block}
The keypoint of the A$^2$F is that it adopts attentive auxiliary feature blocks to utilize all the usable features. Given features $x_0, x_1, ..., x_{i-1}$ from all previous blocks, it is improper to directly fuse with features of the current block because features of different blocks are in different feature spaces. Thus we need to project auxiliary features to a common-space that is suitable to be fused, which prevent features of different space from causing extraodinary imbalance for attention weights. In A$^2$F, 1$\times$1 convolution layer $\mathcal{F}^{i}_{proj}$ is served as such a projection unit. The projected features of the $i_{th}$ auxiliary block $x^{proj}_{i}$ are obtained by
\begin{equation}
   x^{proj}_{i} = \mathcal{F}^{i}_{proj}([x_0, x_1, ..., x_{i-1}]),
\end{equation}
where $[x_0, x_1, ..., x_{i-1}]$ concatenates $x_0, x_1, ..., x_{i-1}$ along the channel. However, different channels of $x^{proj}_{i}$ have different importance when being fused with features of current layer. Therefore, channel attention $\mathcal{F}^{i}_{att}$ is used to learn the importance factor of different channel of $x^{proj}_{i}$. In this way, we get the new features $x^{att}_{i}$ by
\begin{equation}
   x^{att}_{i} = \mathcal{F}^{i}_{att}(x^{proj}_{i})\otimes x^{proj}_{i},
\end{equation}
where $\mathcal{F}^{i}_{att}$ consists of one average pooling layer, one 1$\times$1 convolution layer, one ReLU layer, another 1$\times$1 convolution layer and one sigmoid layer. The symbol $\otimes$ means channel-wise multiplication. The block of WDSR\_A~\cite{WDSR} is adopted to get the features of current layer $x^{res}_{i}$:
\begin{equation}
   x^{res}_{i} = \mathcal{F}^{i}_{res}(x_{i-1}),
\end{equation}
where $\mathcal{F}^{i}_{res}$ consists of one 3$\times$3 convolution layer, one ReLU layer and another 3$\times$3 convolution layer. The output of $i_{th}$ attentive auxiliary feature block $x_{i}$ is given by:
\begin{equation}
   x_{i} = \lambda^{res}_{i}\times x^{res}_{i} + \lambda^{att}_{i}\times x^{att}_{i} + \lambda^{x}_{i}\times x_{i-1},
\end{equation}
where $\lambda^{res}_{i}$, $\lambda^{att}_{i}$ and $\lambda^{x}_{i}$ are feature factors for different features like~\cite{AWSRN}. These feature factors will be learned automatically when training the model. Here we choose additive operation for it can better handle the situation that the $\lambda^{att}_{i}$ of some auxiliary features is 0. If we concat channels directly, there will be some invalid channels which may increase the redundancy of the network. We can also reduce parameters by additive operation sin it does not expand channels.

\section{Experiments}

In this section, we first introduce some common datasets and metrics for evaluation. Then, we describe details of our experiment and analyze the effectiveness of our framework. Finally, we compare our model with state-of-the-art methods both in qualitation and quantitation to demonstrate the superiority of A$^2$F. For more experiments please refer to the supplementary materials.

\subsection{Dataset and Evaluation Metric}
DIV2K dataset~\cite{DIV2K} with 800 training images is used in previous methods~\cite{AWSRN,FC2N} for model training.  When testing the performance of the models, Peak Signal to Noise Ratio (PSNR) and the Structural SIMilarity index (SSIM)~\cite{SSIM} on the Y channel after converting to YCbCr channels are calculated on five benchmark datasets including Set5~\cite{Set5}, Set14~\cite{Set14}, B100~\cite{B100}, Urban100~\cite{Urban100} and Manga109~\cite{Manga109}. We also adopt the LPIPS~\cite{LPIPS} as a perceptual metric to do comparison, which can avoid the situation that over-smoothed images may present a higher PSNR/SSIM when the performances of two methods are similar.

\subsection{Implementation Details}
Similar to AWSRN~\cite{AWSRN}, we design four variants of A$^2$F, denoted as A$^2$F-S, A$^2$F-SD, A$^2$F-M and A$^2$F-L. The channels of $\mathcal{F}^{i}_{res}$ in the attentive auxiliary feature block of A$^2$F-S, A$^2$F-M and A$^2$F-L are set to \{32,128,32\} channels, which means the input, internal and output channel number of $\mathcal{F}^{i}_{res}$ is 32, 128, 32, respectively. The channels of $\mathcal{F}^{i}_{res}$ in the attentive auxiliary feature block of A$^2$F-SD is set to \{16,128,16\}. For the A$^2$F-SD model, we change all of the channels that are setted as 32 in A$^2$F-S, A$^2$F-M, A$^2$F-L to 16. The number of the attentive auxiliary feature blocks of A$^2$F-S, A$^2$F-SD, A$^2$F-M and A$^2$F-L is 4, 8, 12, and 16, respectively. During the training process, typical data augmentation including horizontal flip, rotation and random rotations of $90^o$, $180^o$, $270^o$ are used. The model is trained using Adam algorithm~\cite{adam} with L1 loss. The initial value of $\lambda^{res}_{i}$, $\lambda^{att}_{i}$ and $\lambda^{x}_{i}$ are set to 1. All the code are developed using PyTorch on a machine with an NVIDIA 1080 Ti GPU.

\subsection{Ablation Study}
In this section, we first demonstrate the effectiveness of the proposed auxiliary features. Then, we conduct an ablation experiments to study the effect of essential components of our model and the selection of the kernel for the head component.

\subsubsection{Effect of auxiliary features}

\begin{figure*}[t]
  \centering
  
  \subfigure{
    \label{A2F-s2}
    \includegraphics[width=0.22\textwidth]{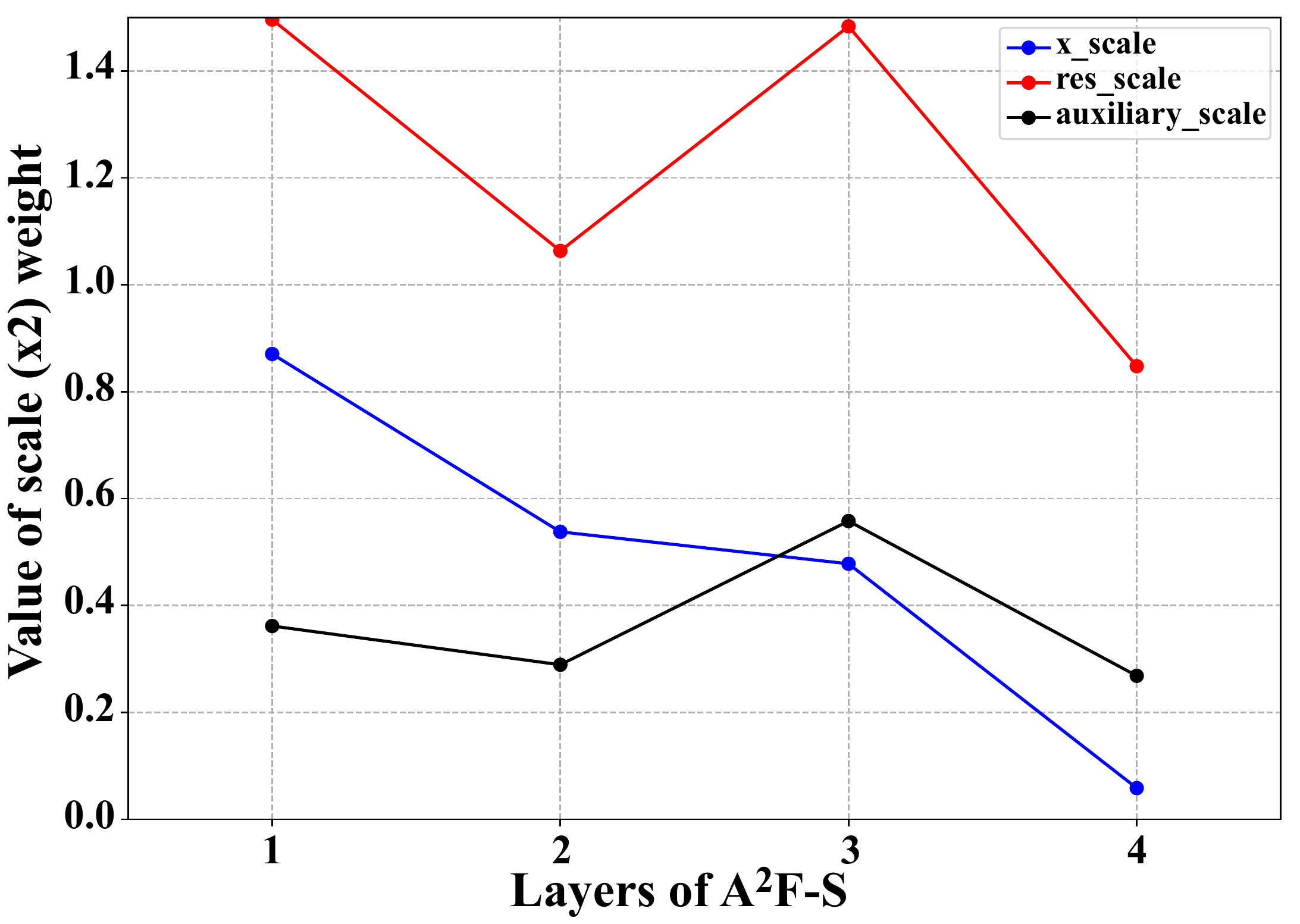}
  }
  \subfigure{
    \label{A2F-s3}
    \includegraphics[width=0.22\textwidth]{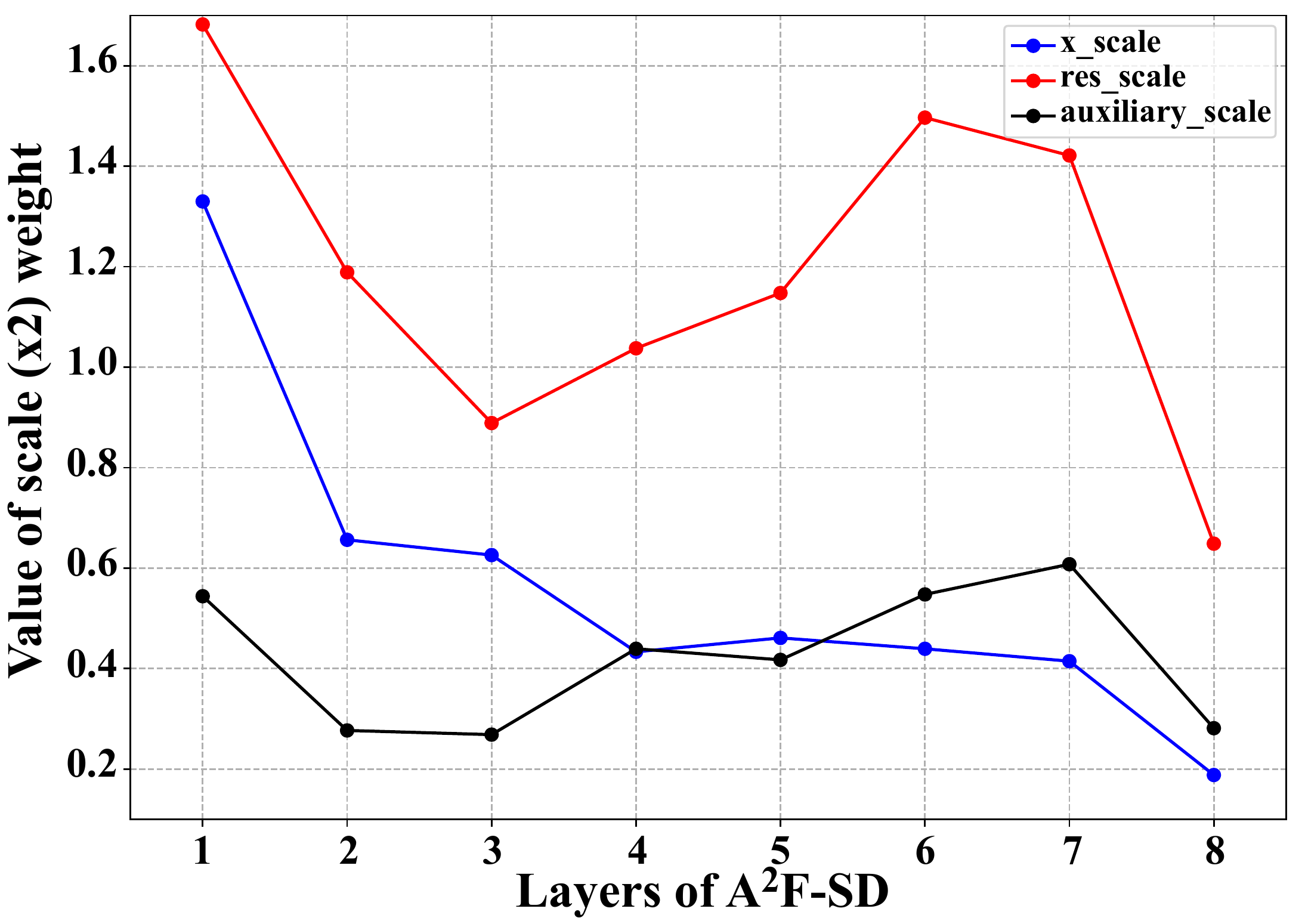}
  }
  \subfigure{
    \label{A2F-s4}
    \includegraphics[width=0.22\textwidth]{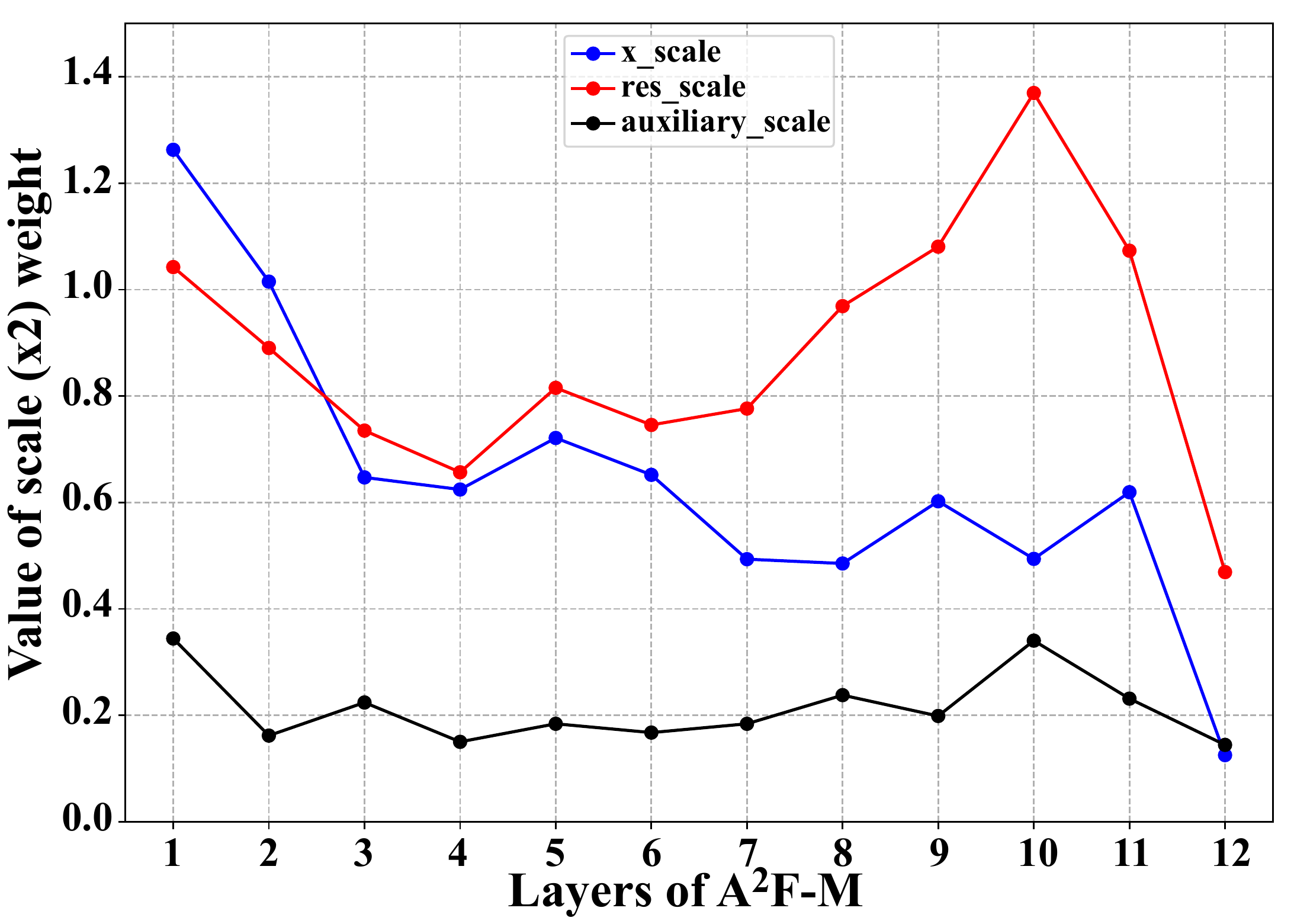}
  }
  \subfigure{
    \label{A2F-sd2}
    \includegraphics[width=0.22\textwidth]{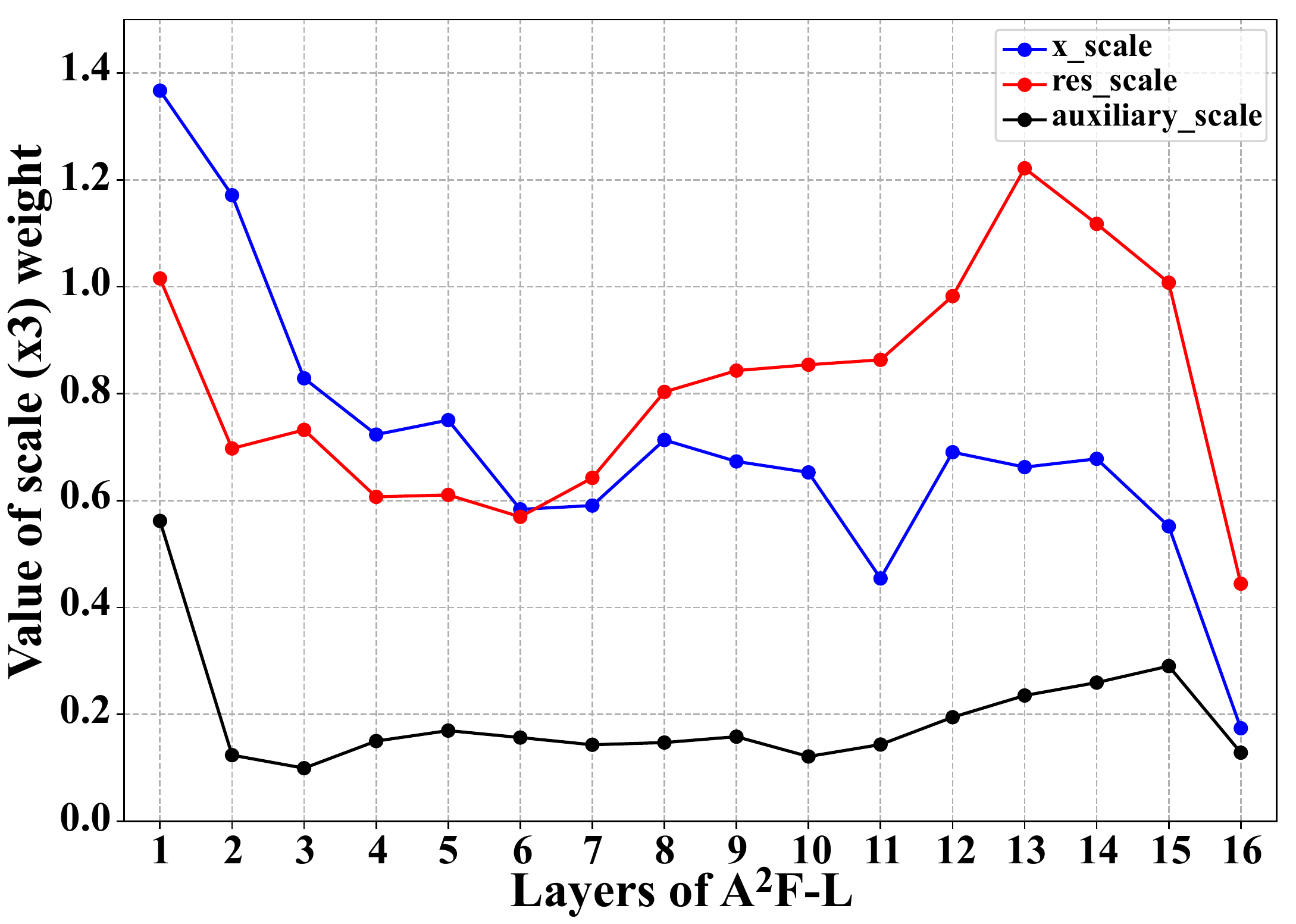}
  }
  
  \subfigure{
    \label{A2F-sd3}
    \includegraphics[width=0.22\textwidth]{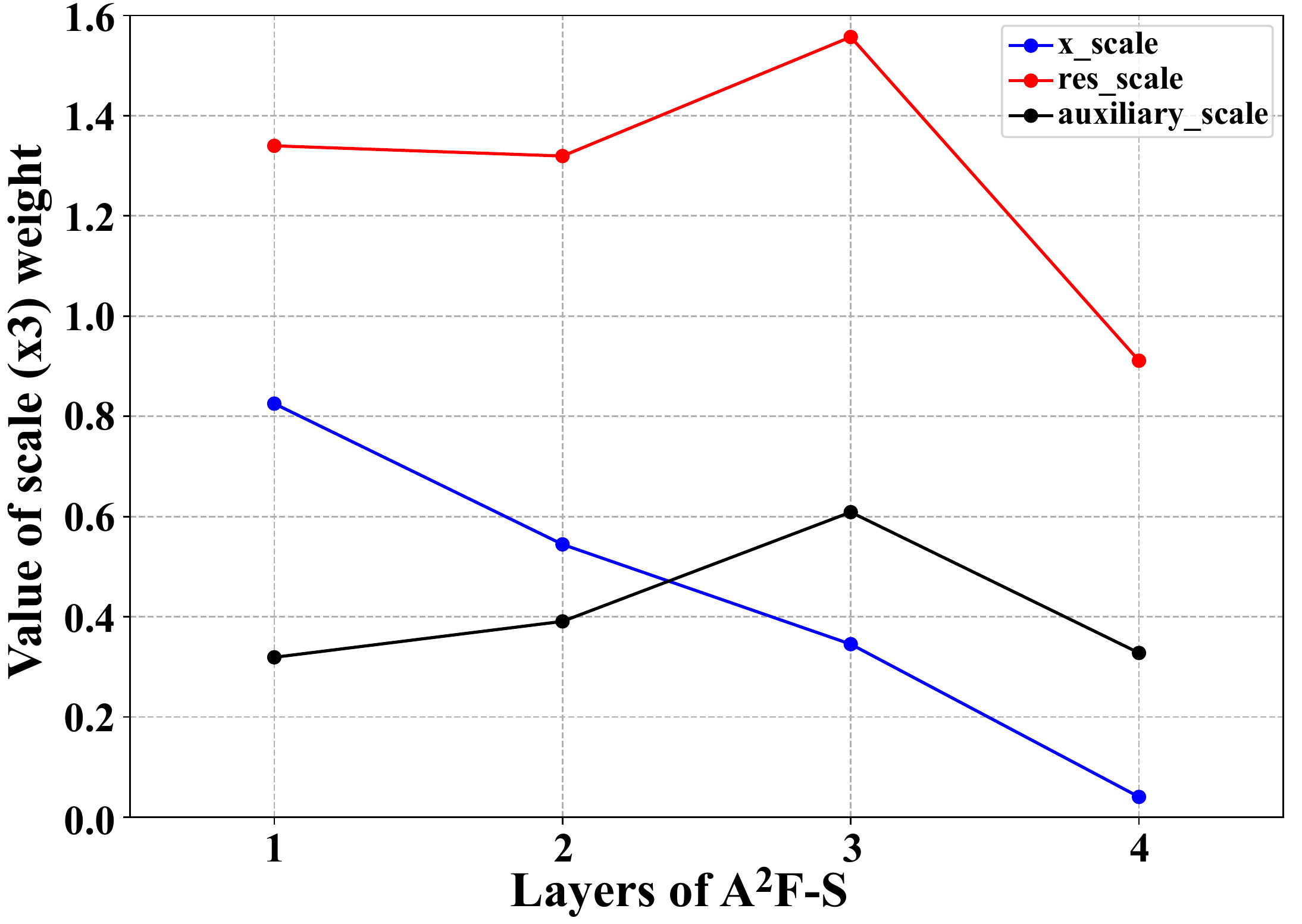}
  }
  \subfigure{
    \label{A2F-sd4}
    \includegraphics[width=0.22\textwidth]{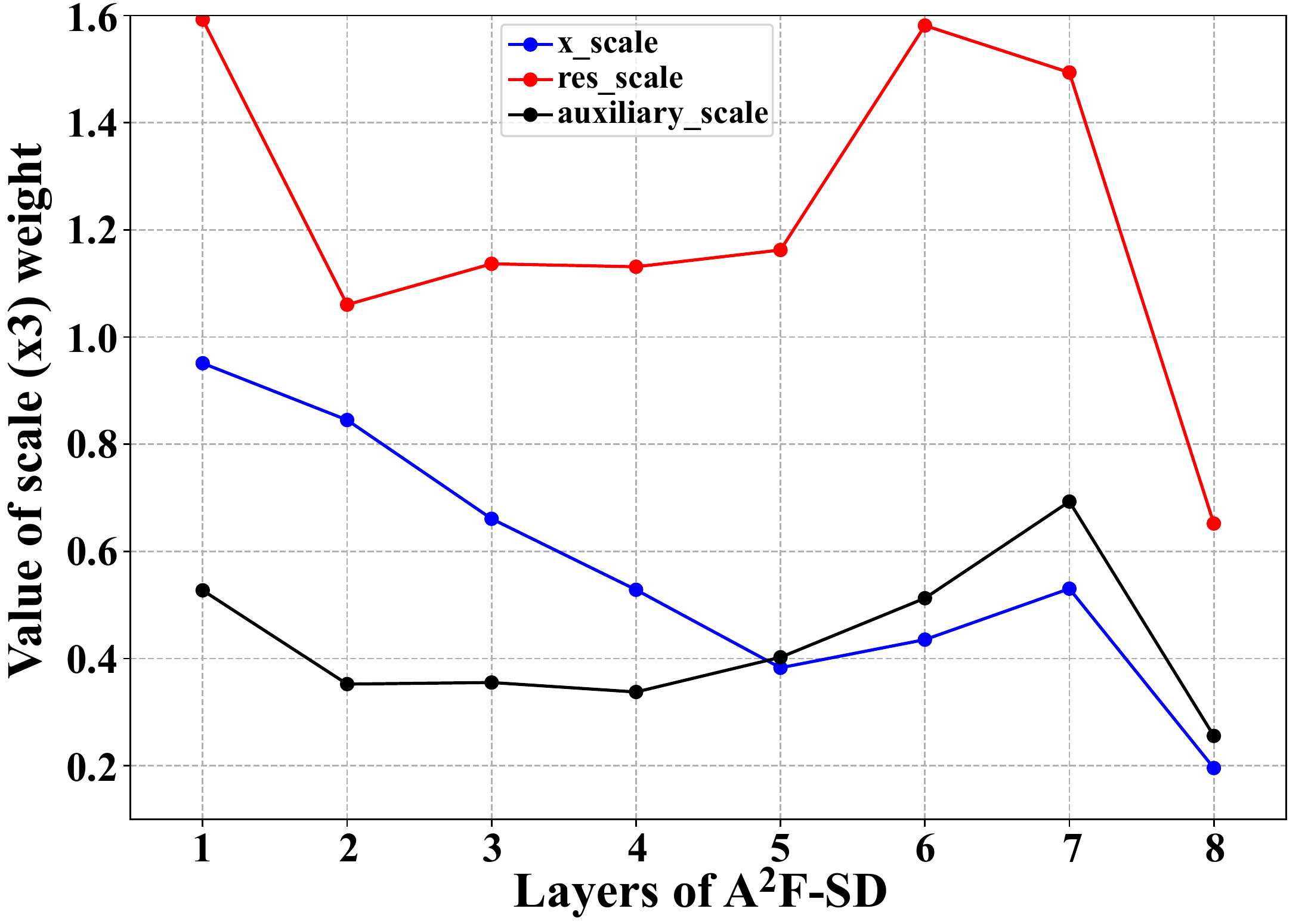}
  }
  \subfigure{
    \label{A2F-m2}
    \includegraphics[width=0.22\textwidth]{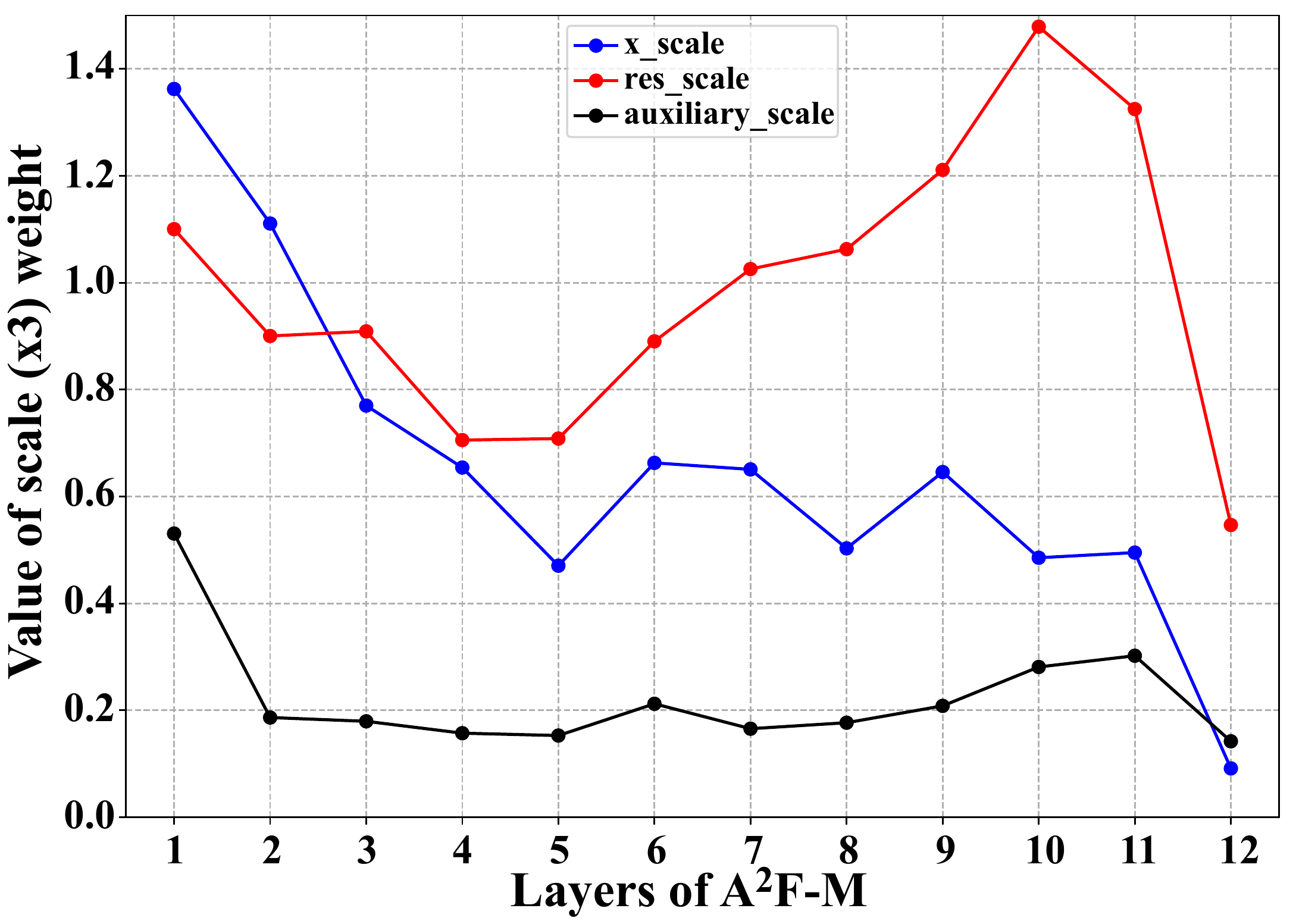}
  }
  \subfigure{
    \label{A2F-m3}
    \includegraphics[width=0.22\textwidth]{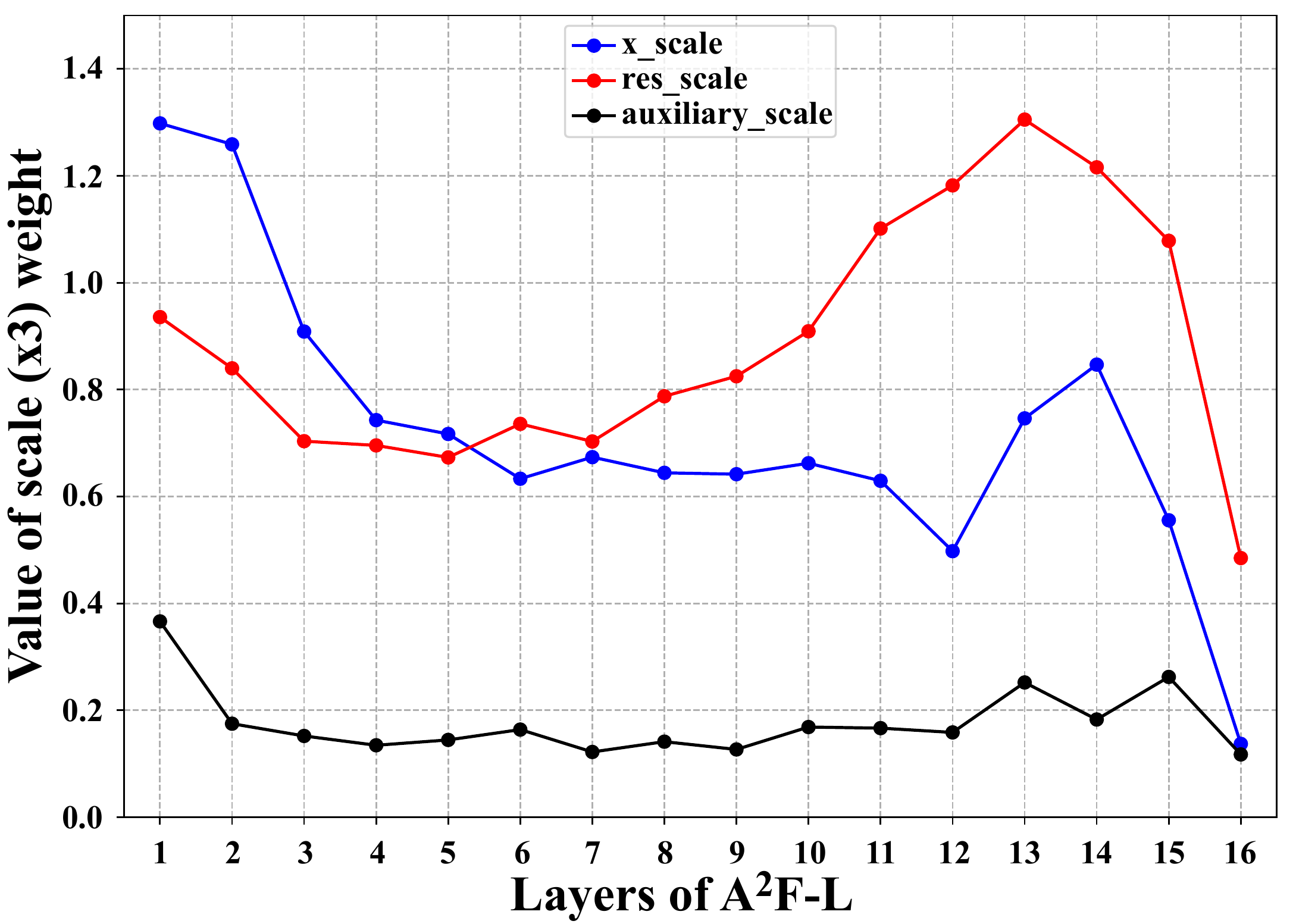}
  }
  
  \subfigure{
    \label{A2F-m4}
    \includegraphics[width=0.22\textwidth]{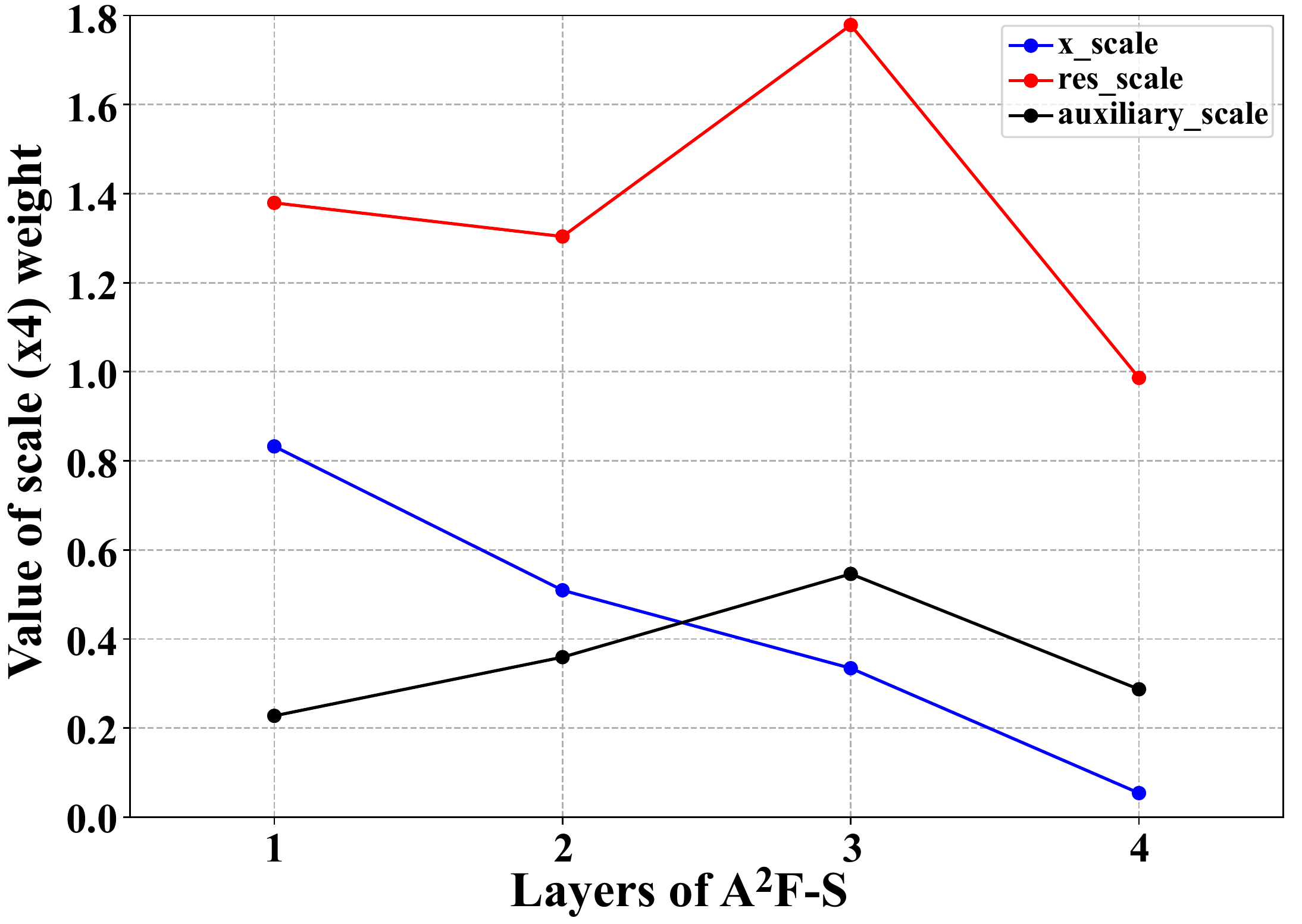}
  }
  \subfigure{
    \label{A2F-l2}
    \includegraphics[width=0.22\textwidth]{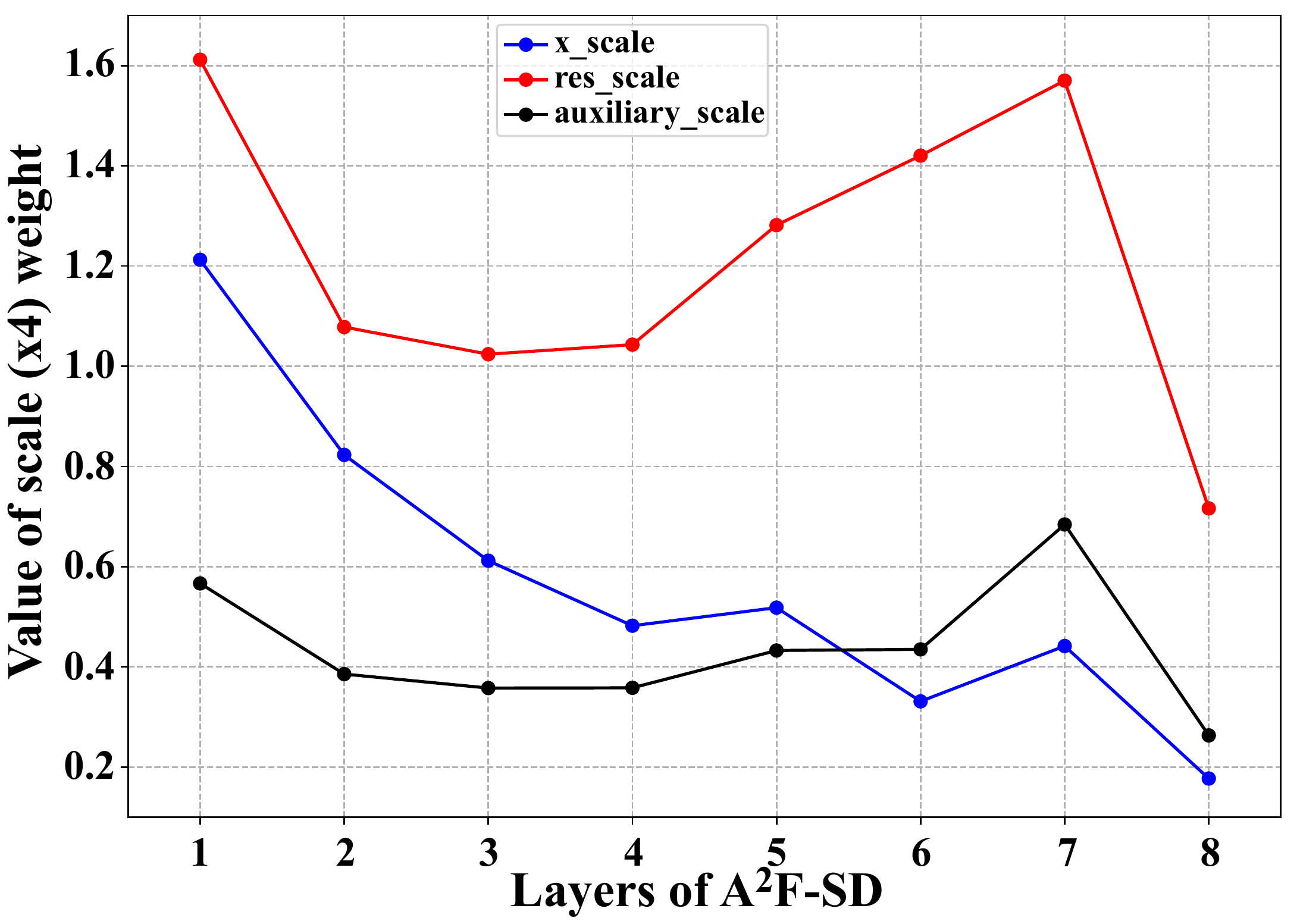}
  }
  \subfigure{
    \label{A2F-l3}
    \includegraphics[width=0.22\textwidth]{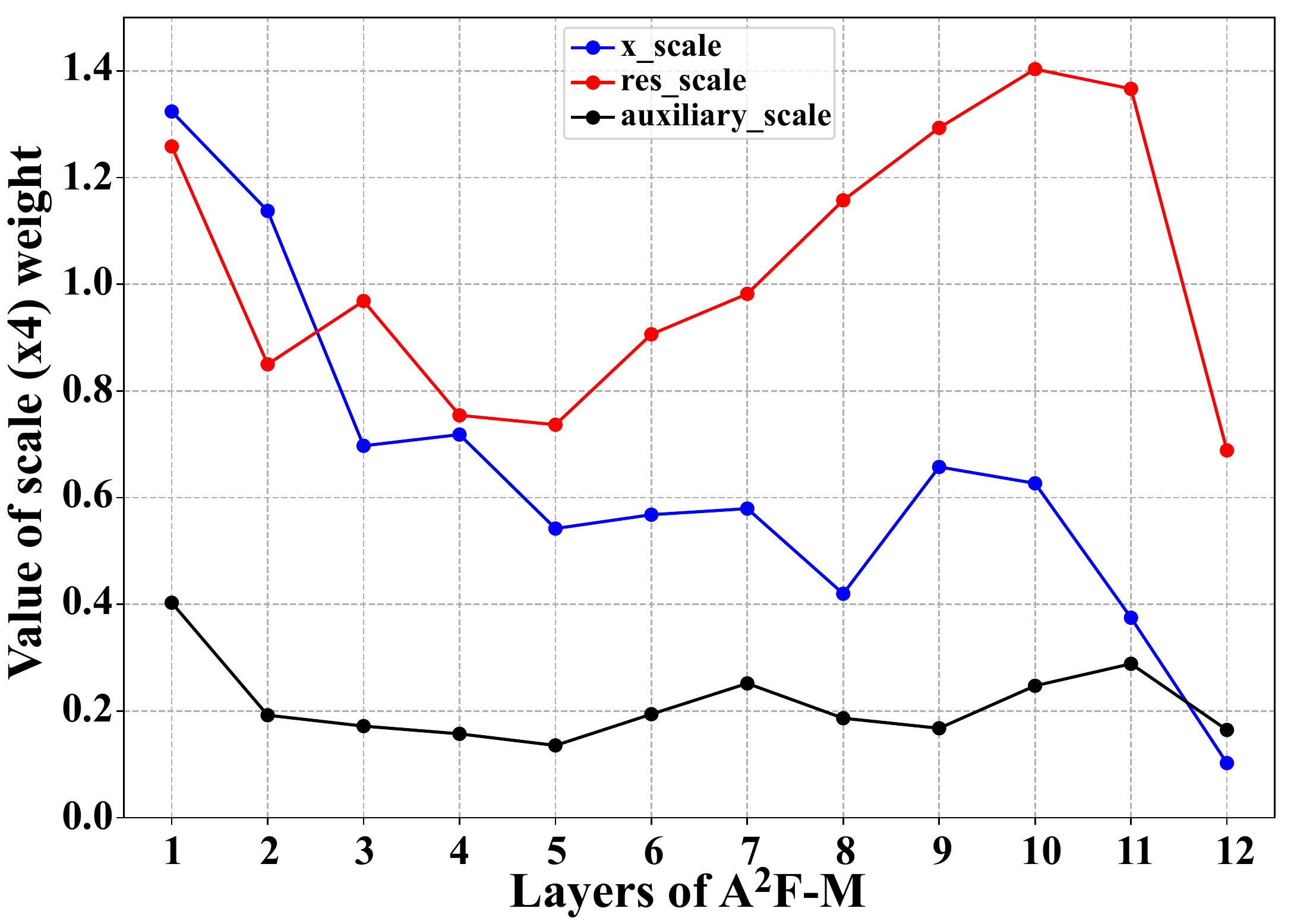}
  }
  \subfigure{
    \label{A2F-l4}
    \includegraphics[width=0.22\textwidth]{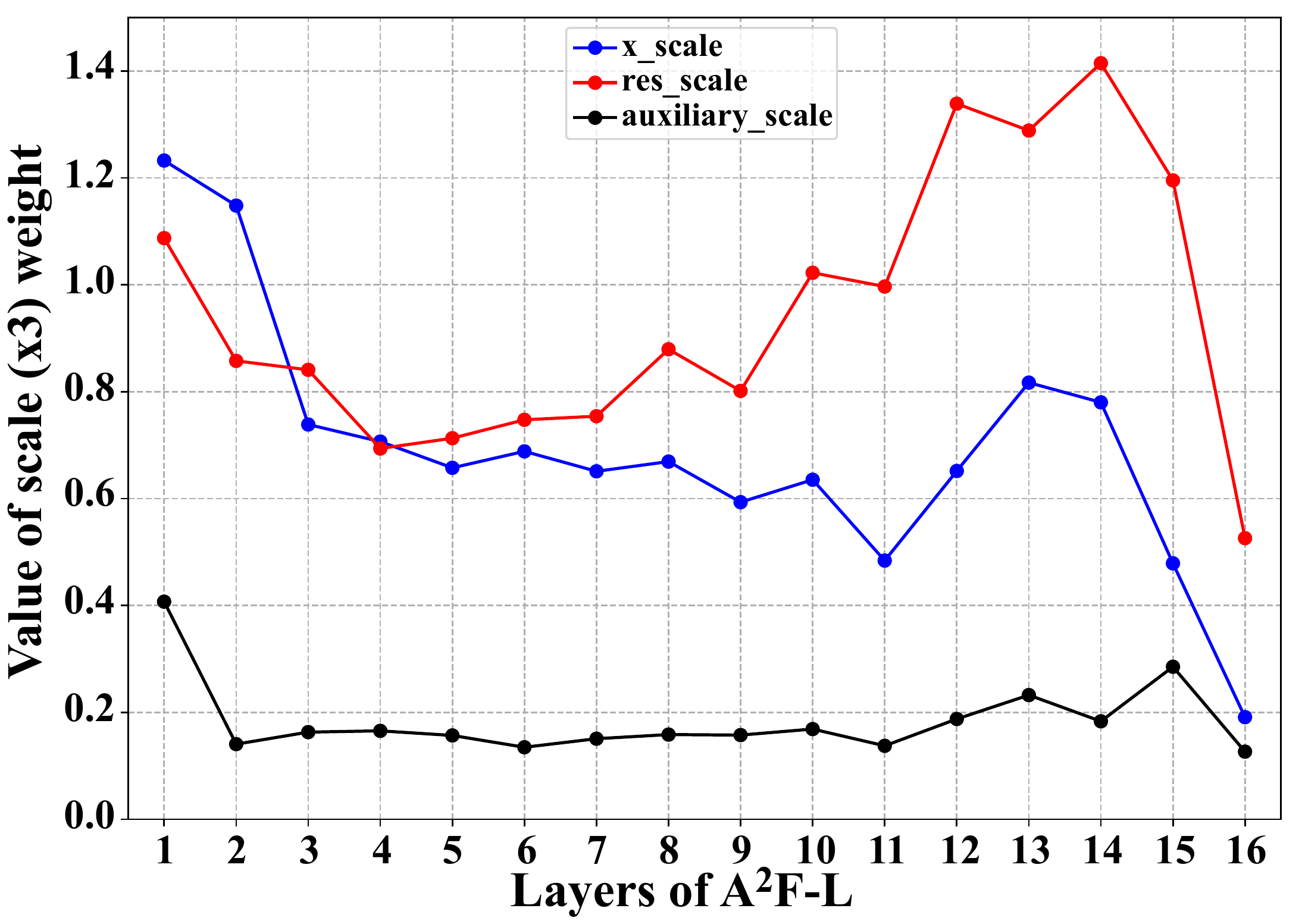}
  }
  
  \caption{The weight of $\lambda^{res}_{i}$ (res\_scale), $\lambda^{att}_{i}$ (auxiliary\_scale) and $\lambda^{x}_{i-1}$ (x\_scale) in different layers. From top to bottom are the results on the $\times2$, $\times3$, $\times4$ tasks. From left to right are the results of models A$^2$F-S, A$^2$F-SD, A$^2$F-M and A$^2$F-L.}
  \label{scaleTable}
\end{figure*}

To show the effect of auxiliary features, we plot the $\lambda^{res}_{i}$, $\lambda^{att}_{i}$ and $\lambda^{x}_{i-1}$ of each layer of each model in Figure~\ref{scaleTable}. As shown in Figure~\ref{scaleTable}, the value of $\lambda^{att}_{i}$ are always bigger than 0.2, which reflects that the auxiliary features always play a certain role in generating the output features of the auxiliary features block. It can also be observed that in all the models of A$^2$F, the weight of  $x^{res}_i$ (i.e. $\lambda^{res}_{i}$) plays the most important role. The weight of $x_{i-1}$ (i.e. $\lambda^{x}_{i-1}$) is usually larger than $\lambda^{att}_{i}$. However, for the more lightweight SISR models (i.e. A$^2$F-S and A$^2$F-SD), $x^{att}_{i}$  becomes more and more important than $x_{i-1}$ (i.e.  $\lambda^{att}_{i}$ becomes more and more larger than $\lambda^{x}_{i-1}$) as the number of layers increases. This reflects that auxiliary features may have great effects on the lightweight SISR models.

\subsubsection{Effect of projection unit and channel attention}

\begin{table*}[tb!]
  \centering
  \caption{Results of ablation study on the projection unit and the channel attention. PSNR is calculated on the super-resolution task with a scale factor of 2. PU means projection unit and CA means channel attention. ``MP'' in the model means more parameters.}
  \resizebox{\textwidth}{13mm}{
  \begin{tabular}{|l|c|c|c|c|c|c|c|c|c|}
    \hline
    \multicolumn{1}{|c|}{Model} & PU & CA & Param & MutiAdds & Set5 & Set14 & B100 & Urban100 & Manga109\\
    \hline
    BASELINE & & & 1190K & 273.9G & 38.04 & 33.69 & 32.20 & 32.20  & 38.66 \\
    BASELINE-MP & & & 1338K & 308.0G & 38.09 & 33.70 & 32.21 & 32.25 & 38.69 \\
    A$^2$F-L-NOCA & $\surd$ &  & 1329K & 306.0G & 38.08 & 33.75 & 32.23 & 32.39 & 38.79 \\
    A$^2$F-L-NOCA-MP & $\surd$ &  & 1368K & 315.1G & 38.09 & 33.77 & 32.23 & 32.35 & 38.79 \\
    A$^2$F-L & $\surd$ & $\surd$ & 1363K & 306.1G & \textbf{38.09} & \textbf{33.78} & \textbf{32.23} & \textbf{32.46} & \textbf{38.95}\\
    \hline
  \end{tabular}}
  
  \label{puAndCa}

\end{table*}

To evaluate the performance of the projection unit and channel attention in the attentive auxiliary feature block, WDSR\_A~\cite{WDSR} with 16 layers is used as the BASELINE model. Then we drop the channel attention in the attentive auxiliary feature block and such model is denoted as A$^2$F-L-NOCA. To further prove the performance gain comes from the proposed attention module, we perform an experiment as follows: we increase the number of parameters of BASELINE and A$^2$F-L-NOCA, and we denote these models as BASELINE-MP and A$^2$F-L-NOCA-MP, where MP means more parameters. Table~\ref{puAndCa} shows that comparing the results of BASELINE, BASELINE-MP and A$^2$F-L-NOCA, we can find that projection unit with auxiliary features can boost the performance on all the datasets. Comparing the results of A$^2$F-L-NOCA, A$^2$F-L-NOCA-MP, A$^2$F-L, it can be found that channel attention in the attentive auxiliary feature block further improves the performance. Thus, we draw the conclusion that the projection unit and channel attention in the auxiliary can both better explore the auxiliary features. In our supplementary materials, we also do this ablation study on a challengeable case (i.e. A$^2$F-S for x4) to show that the good using of auxiliary features is especially important for shallow networks.

\begin{table}[tb!]
  \centering
  \caption{Results of ablation study on different kernel size which is only used for head component. Note that other convolutional kernels are same.}
 
  \resizebox{0.7\textwidth}{12mm}{
    
    \begin{tabular}{|c|c|c|c|c|c|c|}
      \hline
      \multicolumn{7}{|c|}{Convolutional Kernel Selection} \\
      \hline
      \multicolumn{1}{|c|}{Kernel} & Parameters & Set5 & Set14 & B100 & Urban100 & Manga109\\
      \hline
      $1\times1$ & 319.2K  & 32.00 & 28.46 & 27.46 & 25.78  & 30.13 \\
      $\bm{3\times3}$ & \textbf{319.6K} & \textbf{32.06} & \textbf{28.47} & \textbf{27.48} & \textbf{25.80} & \textbf{30.16} \\
      $5\times5$  & 320.4K  & 32.00 & 28.45 & 27.48 & 25.80 & 38.13 \\
      $7\times7$  & 321.6K  & 31.99 & 28.44 & 27.48 & 25.78 & 30.10 \\
      \hline
  \end{tabular}}
  
  \label{kernelsize}

\end{table}

\subsubsection{Kernel selection for $\mathcal{F}_{head}$}
We select different size of kernels in $\mathcal{F}_{head}$ to verify that $1\times1$ conv and large receptive field are not suitable for the head component. From Table~\ref{kernelsize}, we can observe both of them have whittled the performance of the network. This result verifies the reasonability of our head component which has been introduced in section 3.2

\subsection{Comparison with State-of-the-art Methods}
We report an exhaustive comparative evaluation, comparing with several high performance but low parameters and multi-adds operations methods on five datasets, including FSRCNN~\cite{FSRCNN}, DRRN~\cite{DRRN}, FALSR~\cite{FALSR}, CARN~\cite{CARN}, VDSR~\cite{VDSR}, MemNet~\cite{MemNet}, LapSRN~\cite{LapSRN}, AWSRN~\cite{AWSRN}, DRCN~\cite{DRCN}, MSRN~\cite{MSRN}, SRMDNF~\cite{SRMDNF}, SelNet~\cite{SelNet}, IDN~\cite{IDN}, SRFBN-S~\cite{SRFBN} and so on. Note that we do not consider methods that have significant performance such as RDN~\cite{RDN}, RCAN~\cite{RCAN}, EDSR~\cite{EDSR} for they have nearly even more than 10M parameters. It is unrealistic to apply the method in real-world application though they have higher PSNR. But we provide a supplementary material to compare with these non-lightweight SOTAs. To ensure that parameters of different methods are at the same magnitude, we divide the comparison experiment on a single scale into multi-group according to different parameters. All methods including ours have been evaluated on $\times 2$, $\times 3$, $\times 4$.

\begin{figure*}[!t]
  \centering
  \includegraphics[width=0.96\textwidth]{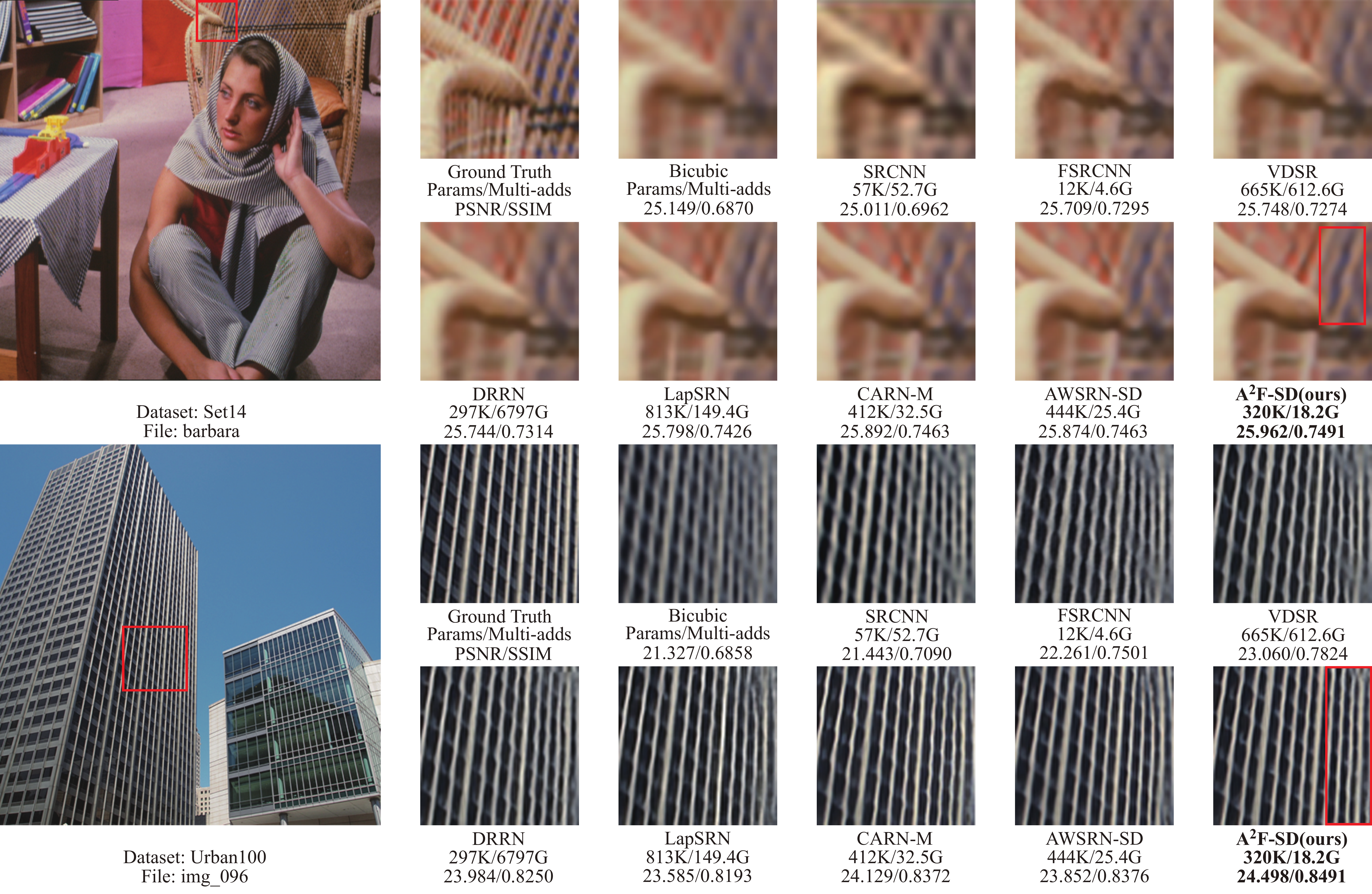}
  \caption{Qualitative comparison over datasets for scale $\times4$. The red rectangle indicates the area of interest for zooming. Comparison for other two datasets can be seen supplementary material.}
  \label{Quality comparison of different models.}
\end{figure*}

\subsubsection{Qualitative comparison}
Qualitative comparison is shown in Figure~\ref{Quality comparison of different models.}. We choose methods whose parameters are less than 1000k since we think high efficiency (low parameters) is essential. We can see that our method A$^2$F-SD achieves better performance than others, which is represented through recovering more high-frequency information for the entire image. For the image barbara in Set14 (row 1 in Figure~\ref{Quality comparison of different models.}), our method performs a clear difference between the blue area and the apricot area on the right top corner of the image. Compared with AWSRN-SD which is the second method in our table, our model removes more blur and constructs more regular texture on the right top corner of the image img096 of Urban100. We own this advantage to the sufficient using of auxiliary features of previous layers which incorporate multi-scale features in different convolution progress that might contain abundant multi-frequency information. While the attention mechanism conduces to the adaptive selection of different frequency among various layers.

\subsubsection{Quantitative comparison}
Table~\ref{quantityresults} shows the detailed comparison results. Our models obtain a great trade-off between performance and parameters. In particular, when the number of parameters is less than 1000K, our model achieves the best result for arbitrary scales on each dataset among all of the algorithms. A$^2$F-SD, which only has about 300K parameters, even shows better performance on a variety of datasets compared to DRCN that has nearly 1800K parameters. This proves the tremendous potential of A$^2$F for real-world application. The high efficiency of A$^2$F comes from the mechnism of sufficient fusion of former layers feature via the proposed attention scheme. Because we adopt 1$\times$1 Conv and channel attention to select the appropriate features of former layers for fusing, which can help to reduce the number of layers in the network without sacrificing good performance. When the number of parameters is more than 1000K, A$^2$F-L model also performs a SOTA result on the whole, although worse in some cases slightly. It is due to that they combine all features of former layers without considering whether they are useful, which cause a reduction to performance. While compared to AWSRN-M and AWSRN, A$^2$F-M model has more advantage in trade-off since it has comparable PSNR and SSIM but only 1010K parameters that account for 63\%, 80\% of AWSRN and AWSRN-M, respectively.

\begin{table}[!t]
  \centering
  \caption{Running time comparison with $\times4$ scale on Urban100 dataset. All of them are evaluated on the same mechine.}
  \smallskip
  \resizebox{0.7\textwidth}{!}{
    \begin{tabular}{|c|c|c|c|c|}
      \hline
      Model & Params & Multi-Adds& Running time(s) & PSNR \\
      \hline
      RCAN~\cite{RCAN}  & 15590K &919.9G& 0.8746  &  26.82  \\
      EDSR~\cite{EDSR} & 43090K &2896.3G& 0.3564 &26.64 \\
      D-DBPN~\cite{DBPN} & 10430K &685.7G& 0.4174 & 26.38 \\
      SRFBN~\cite{SRFBN} & 3631K &1128.7G& 0.4291  &  26.60  \\
      SRFBN-S~\cite{SRFBN} & 483K & 132.5G &0.0956 & 25.71  \\
      VDSR~\cite{VDSR} & 665K & 612.6G & 0.1165 & 25.18 \\
      CARN-M~\cite{CARN} & 412K & 32.5G & 0.0326 & 25.62 \\
      \hline
      \textbf{A}$\bm{^2}$\textbf{F-SD} & \textbf{320K} &\textbf{18.2G}& \textbf{0.0145} &  \textbf{25.80} \\
      \textbf{A}$\bm{^2}$\textbf{F-L}  & \textbf{1374K} &\textbf{77.2G}& \textbf{0.0324} &  \textbf{26.32} \\

      \hline
  \end{tabular}}
  
  \label{runningtime}
 
\end{table}

\subsection{Running Time and GFLOPS}
We compare our model A$^2$F-SD and A$^2$F-L with other methods (both lightweight ~\cite{VDSR,SRFBN,CARN} and non-lightweight~\cite{RCAN,EDSR,DBPN}) in running time to verify the high efficiency of our work in Table~\ref{runningtime}. Like~\cite{SRFBN}, we evaluate our method on a same machine with four NVIDIA 1080Ti GPUs and 3.6GHz Intel i7 CPU. All of the codes are official implementation. To be fair, we only use a single NVIDIA 1080Ti GPU for evaluation, and only contain codes that are necessary for testing an image, which means operations of saving images, saving models, opening log files, appending extra datas and so on are removed from the timing program. 

To reduce the accidental error, we evaluate each method for four times on each GPU and calculate the avarage time as the final running time for a method. Table~\ref{runningtime} shows that our models represent a significant surpass on running time for an image compared with other methods, even our A$^2$F-L model is three times faster than SRFBN-S~\cite{SRFBN} which has only 483K parameters with 25.71 PSNR. All of our models are highly efficient and keep being less comparable with RCAN~\cite{RCAN} which are 60 and 27 times slower than our A$^2$F-SD, and A$^2$F-L model, respectively. This comparison result reflects that our method gets the tremendous trade-off between performance and running time and is the best choice for realistic applications.

We also calculate the GFLOPs based on the input size of $32\times32$ for several methods that can be comparable with A$^2$F in Table~\ref{LPIPS}. We actually get high performance with lower GFLOPs both for our large and small models.

\begin{table}[tb!]
  \centering
  \caption{The perceptual metric LPIPS on five datasets for scale x4. The lower is better. We only choose methods that can be comparable with A$^2$F. All of the output SR images are provided officially.}
  \resizebox{0.7\textwidth}{16mm}{
    \begin{tabular}{|c|c|c|c|c|c|c|c|}
      \hline
      Methods &Params&GFLOPs& Set5 & Set14 & B100 & Urban100 & Manga109 \\
      \hline
      AWSRN~\cite{AWSRN} & 1587K & 1.620G  &    0.1747&   0.2853 &  {\color{red}0.3692} &  0.2198 &  0.1058 \\
      AWSRN-SD~\cite{AWSRN} & 444K& - & 0.1779 &   0.2917 &   0.3838 &   0.2468&  0.1168 \\
      CARN~\cite{CARN}   &1592K& 1.620G & 0.1761 &  0.2893 &  0.3799 &   0.2363 & - \\
      CARN-M~\cite{CARN} &412K &0.445G&  0.1777 &  0.2938 &  0.3850  &  0.2524 & - \\
      SRFBN-S~\cite{SRFBN} & 483K& 0.323G & 0.1776  & 0.2938 & 0.3861 & 0.2554 & 0.1396 \\
      IMDN~\cite{IMDN}& 715K& 0.729G& 0.1743  &  0.2901  & 0.3740 & 0.2350  & 0.1330 \\
      \hline
      \textbf{A$^2$F-SD} & 320K& 0.321G&{\color{red}0.1731}& 0.2870 &0.3761 & 0.2375 &  0.1112 \\
      \textbf{A$^2$F-L} & 1374K&1.370G &  0.1733 & {\color{red}0.2846} & 0.3698 & {\color{red}0.2194} & {\color{red}0.1056} \\
      
      \hline
      
  \end{tabular}}

  \label{LPIPS}
\end{table}

\subsection{Perceptual Metric}
Perceptual metric can better reflect the judgment of image quality. In this paper, LPIPS~\cite{LPIPS} is chosen as the perceptual metric. From Table~\ref{LPIPS}, our proposed model obtains superior results with high efficiency in most cases, which shows their ability of generating more realistic images.

\section{Conclusion}
In this paper, we propose a lightweight single-image super-resolution network called A$^2$F which adopts attentive auxiliary feature blocks to efficiently and sufficiently utilize auxiliary features. Quantitive experiment results demonstrate that auxiliary features with projection unit and channel attention can achieve higher PSNR and SSIM as well as perceptual metric LPIPS with less running time on various datasets. Qualitative experiment results reflect that auxiliary features can give the predicted image more high-frequency information, thus making the models achieve better performance. The A$^2$F model with attentive auxiliary feature block is easy to implement and achieves great performance when the number of parameters is less than 320K and the multi-adds are less than 75G, which shows that it has great potential to be deployed in practical applications with limited computation resources. In the future, we will investigate more measures to better fuse auxiliary features.

\subsubsection{Acknowledgement}
This work was supported in part by the National Key Research and Development Program of China under Grant 2018YFB1305002, in part by the National Natural Science Foundation of China under Grant 61773414, and Grant 61972250, in part by the Key Research and Development Program of Guangzhou under Grant 202007050002, and Grant 202007050004.

\begin{table*}[!t]
  \centering
  \caption{Evaluation on five datasets by scale $\times2$, $\times3$, $\times4$. {\color{red}Red} and {\color{blue}blue} imply the best and second best result in a group, respectively.}
  \resizebox{\textwidth}{90mm}{
    \begin{tabular}{|c|c|l|c|c|c|c|c|c|c|}
      \hline
      Scale&Size Scope&\multicolumn{1}{c|}{Model}&Param&MutiAdds&Set5&Set14&B100&Urban100&Manga109\\
      \hline
      \multirow{26}{*}{x2}&\multirow{11}{*}{$<5\times10^2K$}& FSRCNN &12K&6G&37.00/0.9558&32.63/0.9088 & 31.53/0.8920 & 29.88/0.9020 & 36.67/0.9694 \\
      && SRCNN& 57K  & 52.7G   & 36.66/0.9542   & 32.42/0.9063 & 31.36/0.8879 & 29.50/0.8946 & 35.74/0.9661 \\
      && DRRN         & 297K & 6797G   & 37.74/0.9591   & 33.23/0.9136 & 32.05/0.8973 & 31.23/0.9188 & 37.92/0.9760 \\
      && \textbf{A$^2$F-SD(ours)} & \textbf{313k}& \textbf{71.2G}& {\color{red} \textbf{37.91/0.9602}}& {\color{red}\textbf{33.45/0.9164}}&{\color{red} \textbf{32.08/0.8986}}&{\color{red} \textbf{31.79/0.9246}}&{\color{red} \textbf{38.52/0.9767 }}\\
      && \textbf{A$^2$F-S(ours)}  & \textbf{320k} & \textbf{71.7G}   & \textbf{37.79/0.9597}& \textbf{33.32/0.9152} & \textbf{31.99/0.8972}& \textbf{31.44/0.9211} & \textbf{38.11/0.9757}\\
      && FALSR-B      & 326K & 74.7G   & 37.61/0.9585 & 33.29/0.9143 & 31.97/0.8967 & 31.28/0.9191 & -            \\
      && AWSRN-SD     & 348K & 79.6G   & {\color{blue}37.86/0.9600 }  & {\color{blue}33.41/0.9161} & {\color{blue}32.07/0.8984} & {\color{blue}31.67/0.9237} & {\color{blue}38.20/0.9762 }\\
      && AWSRN-S      & 397K & 91.2G   & 37.75/0.9596   & 33.31/0.9151 & 32.00/0.8974 & 31.39/0.9207 & 37.90/0.9755 \\
      && FALSR-C      & 408K & 93.7G   & 37.66/0.9586 & 33.26/0.9140 & 31.96/0.8965 & 31.24/0.9187 & -            \\
      && CARN-M       & 412K & 91.2G   & 37.53/0.9583   & 33.26/0.9141 & 31.92/0.8960 & 31.23/0.9193 & -            \\ 
      && SRFBN-S      & 483K&  -      & 37.78/0.9597  & 33.35/0.9156 & 32.00/0.8970 & 31.41/0.9207 & 38.06/0.9757   \\  \cline{2-10}
      &\multirow{6}{*}{$<10^3K$}& IDN & 552K & - & 37.83/0.9600 & 33.30/0.9148 &  32.08/0.8985 & 31.27/0.9196 & - \\
      && VDSR         & 665K & 612.6G  & 37.53/0.9587   & 33.03/0.9124 & 31.90/0.8960 & 30.76/0.9140 & 37.22/0.9729 \\
      && MemNet       & 677K & 2662.4G & 37.78/0.9597 & 33.28/0.9142 & {\color{blue}32.08/0.8978} & {\color{blue}31.31/0.9195} & -            \\
      && LapSRN       & 813K & 29.9G   & 37.52/0.9590 & 33.08/0.9130 & 31.80/0.8950 & 30.41/0.9100 & {\color{blue}37.27/0.9740} \\
      && SelNet       & 974K & 225.7G  & {\color{blue}37.89/0.9598}   & {\color{blue} 33.61/0.9160} & 32.08/0.8984 & -            & -            \\
      && \textbf{A$^2$F-M(ours)}  & \textbf{999k} & \textbf{224.2G}  & {\color{red} \textbf{38.04/0.9607}}& {\color{red} \textbf{33.67/0.9184}}& {\color{red} \textbf{32.18/0.8996}}& {\color{red} \textbf{32.27/0.9294} }       & {\color{red} \textbf{38.87/0.9774}}\\        \cline{2-10}
      &\multirow{8}{*}{$<2\times10^3K$}&FALSR-A& 1021K&234.7G & 37.82/0.9595 & 33.55/0.9168 & 32.12/0.8987   & 31.93/0.9256 & -  \\
      && MoreMNAS-A  & 1039K & 238.6G  & 37.63/0.9584   & 33.23/0.9138 & 31.95/0.8961   & 31.24/0.9187 & -              \\
      && AWSRN-M     & 1063K & 244.1G  & 38.04/0.9605   & 33.66/0.9181 & 32.21/0.9000   & 32.23/0.9294 & 38.66/0.9772   \\
      && \textbf{A$^2$F-L(ours)} & \textbf{1363k} & \textbf{306.1G}  & {\color{blue}\textbf{38.09/0.9607}}          & {\color{red} \textbf{33.78/0.9192}}        & {\color{blue} \textbf{32.23/0.9002}}          & {\color{blue} \textbf{32.46/0.9313}}        & {\color{red} \textbf{38.95/0.9772}}          \\
      && AWSRN       & 1397K & 320.5G  & {\color{red} 38.11/0.9608 }& {\color{blue} 33.78/0.9189 }& {\color{red} 32.26/0.9006}   & {\color{red} 32.49/0.9316 }&{\color{blue}  38.87/0.9776  } \\
      && SRMDNF      & 1513K & 347.7G  & 37.79/0.9600   & 33.32/0.9150 & 32.05/0.8980   & 31.33/0.9200 & -              \\
      && CARN        & 1592K & 222.8G  & 37.76/0.9590   & 33.52/0.9166 & 32.09/0.8978   & 31.92/0.9256 & -              \\
      && DRCN        & 1774K & 17974G  & 37.63/0.9588   & 33.04/0.9118 & 31.85/0.8942   & 30.75/0.9133 & 37.63/0.9723   \\\cline{2-10}
      &\multirow{1}{*}{$<5\times10^3K$}& MSRN        & 5930K & 1365.4G & 38.08/0.9607   & 33.70/0.9186 & 32.23/0.9002   & 32.29/0.9303 & 38.69/0.9772 \\
      \hline
      \multirow{21}{*}{x3}&\multirow{12}{*}{$<10^3K$}& FSRCNN       & 12K   & 5G      & 33.16/0.9140   & 29.43/0.8242   & 28.53/0.7910   & 26.43/0.8080 & 30.98/0.9212   \\
      && SRCNN        & 57K   & 52.7G   & 32.75/0.9090   & 29.28/0.8209   & 28.41/0.7863   & 26.24/0.7989 & 30.59/0.9107   \\
      && DRRN         & 297K  & 6797G   & 34.03/0.9244   & 29.96/0.8349   & 28.95/0.8004   & 27.53/0.8378 & 32.74/0.9390   \\
      && \textbf{A$^2$F-SD(ours)} & \textbf{316k}  & \textbf{31.9G}   &{\color{red}  \textbf{34.23/0.9259}}          & {\color{red} \textbf{30.22/0.8395}}          & {\color{red} \textbf{29.01/0.8028}}          & {\color{red} \textbf{27.91/0.8465}}        & {\color{red} \textbf{33.29/0.9424}}          \\
      && \textbf{A$^2$F-S(ours)}  & \textbf{324k}  & \textbf{32.3G}   & \textbf{34.06/0.9241}          & \textbf{30.08/0.8370}          & \textbf{28.92/0.8006}          & \textbf{27.57/0.8392}        & \textbf{32.86/0.9394}          \\
      && AWSRN-SD     & 388K  & 39.5G   & {\color{blue} 34.18/0.9273}   & {\color{blue} 30.21/0.8398}   & {\color{blue} 28.99/0.8027}   & {\color{blue} 27.80/0.8444} & {\color{blue} 33.13/0.9416}   \\
      && CARN-M       & 412K  & 46.1G   & 33.99/0.9236   & 30.08/0.8367   & 28.91/0.8000   & 27.55/0.8385 & -              \\
      && AWSRN-S      & 477K  & 48.6G   & 34.02/0.9240   & 30.09/0.8376   & 28.92/0.8009   & 27.57/0.8391 & 32.82/0.9393   \\
      && SRFBN-S      & 483K& - &  34.20/0.9255    &30.10/0.8372& 28.96/0.8010 & 27.66/0.8415 & 33.02/0.9404   \\
      && IDN          & 552K  & -       & 34.11/0.9253   & 29.99/0.8354   & 28.95/0.8013   & 27.42/0.8359 & -              \\
      && VDSR         & 665K  & 612.6G  & 33.66/0.9213   & 29.77/0.8314   & 28.82/0.7976   & 27.14/0.8279 & 32.01/0.9310   \\
      && MemNet       & 677K  & 2662.4G & 34.09/0.9248   & 30.00/0.8350   & 28.96/0.8001   & 27.56/0.8376 & -              \\ \cline{2-10}
      &\multirow{8}{*}{$<2\times10^3K$}  & \textbf{A$^2$F-M(ours)}  & \textbf{1003k} & \textbf{100.0G}  & \textbf{34.50/0.9278}          & {\color{blue}\textbf{30.39/0.8427}}        & \textbf{29.11/0.8054}          & \textbf{28.28/0.8546}       & \textbf{33.66/0.9453}          \\
      && AWSRN-M      & 1143K & 116.6G  & 34.42/0.9275   & 30.32/0.8419   & 29.13/0.8059   & 28.26/0.8545 & 33.64/0.9450   \\
      && SelNet       & 1159K & 120G    & 34.27/0.9257   & 30.30/0.8399   & 28.97/0.8025   & -            & -              \\
      && \textbf{A$^2$F-L(ours)}  & \textbf{1367k} & \textbf{136.3G}  & {\color{red} \textbf{34.54/0.9283}}         & {\color{red} \textbf{30.41/0.8436}}          & {\color{blue}\textbf{29.14/0.8062}}         & {\color{blue} \textbf{28.40/0.8574}}        & {\color{blue}\textbf{33.83/0.9463}}         \\
      && AWSRN        & 1476K & 150.6G  & {\color{blue}34.52/0.9281}  & 30.38/0.8426   & {\color{red}29.16/0.8069}   & {\color{red}28.42/0.8580} & {\color{red}33.85/0.9463}  \\
      && SRMDNF       & 1530K & 156.3G  & 34.12/0.9250   & 30.04/0.8370   & 28.97/0.8030   & 27.57/0.8400 & -              \\
      && CARN         & 1592K & 118.8G  & 34.29/0.9255   & 30.29/0.8407   & 29.06/0.8034   & 28.06/0.8493 & -              \\
      && DRCN         & 1774K & 17974G  & 33.82/0.9226   & 29.76/0.8311   & 28.80/0.7963   & 27.15/0.8276 & 32.31/0.9328   \\ \cline{2-10}
      &\multirow{1}{*}{$<10^4K$}& MSRN         & 6114K & 625.7G  & 34.46/0.9278   & 30.41/0.8437   & 29.15/0.8064   & 28.33/0.8561 & 33.67/0.9456   \\
      \hline
      \multirow{23}{*}{x4}&\multirow{13}{*}{$<10^3K$}& FSRCNN       & 12K   & 4.6G    & 30.71/0.8657   & 27.59/0.7535   & 26.98/0.7150   & 24.62/0.7280 & 27.90/0.8517   \\
      && SRCNN        & 57K   & 52.7G   & 30.48/0.8628   & 27.49/0.7503   & 26.90/0.7101   & 24.52/0.7221 & 27.66/0.8505   \\
      && DRRN         & 297K  & 6797G   & 31.68/0.8888   & 28.21/0.7720   & 27.38/0.7284   & 25.44/0.7638 & 29.46/0.8960   \\
      && \textbf{A$^2$F-SD(ours)} & \textbf{320k}  & \textbf{18.2G}   & {\color{red} \textbf{32.06/0.8928}}          & {\color{red} \textbf{28.47/0.7790}}          & {\color{red} \textbf{27.48/0.7373}}          & {\color{red} \textbf{25.80/0.7767}}         & {\color{red} \textbf{30.16/0.9038}}          \\
      && \textbf{A$^2$F-S(ours)}  & \textbf{331k}  & \textbf{18.6G}   & \textbf{31.87/0.8900}          & \textbf{28.36/0.7760}          & \textbf{27.41/0.7305}          & \textbf{25.58/0.7685}        & \textbf{29.77/0.8987}          \\
      && CARN-M       & 412K  & 32.5G   & 31.92/0.8903   & 28.42/0.7762   & 27.44/0.7304   & 25.62/0.7694 & -              \\
      && AWSRN-SD     & 444K  & 25.4G   & {\color{blue} 31.98/0.8921}   & {\color{blue} 28.46/0.7786}   & {\color{blue} 27.48/0.7368}   & {\color{blue} 25.74/0.7746} & {\color{blue} 30.09/0.9024}   \\
      && SRFBN-S      & 483K& 132.5G   & 31.98/0.8923  & 28.45/0.7779 & 27.44/0.7313 & 25.71/0.7719 &  29.91/0.9008  \\
      && IDN          & 552K  & -       & 31.82/0.8903   & 28.25/0.7730   & 27.41/0.7297   & 25.41/0.7632 & -              \\
      && AWSRN-S      & 588K  & 37.7G   & 31.77/0.8893   & 28.35/0.7761   & 27.41/0.7304   & 25.56/0.7678 & 29.74/0.8982   \\
      && VDSR         & 665K  & 612.6G  & 31.35/0.8838   & 28.01/0.7674   & 27.29/0.7251   & 25.18/0.7524 & 28.83/0.8809   \\
      && MemNet       & 677K  & 2662.4G & 31.74/0.8893   & 28.26/0.7723   & 27.40/0.7281   & 25.50/0.7630 & -              \\
      && LapSRN       & 813K  & 149.4G  & 31.54/0.8850   & 28.19/0.7720   & 27.32/0.7280   & 25.21/0.7560 & 29.09/0.8845   \\ \cline{2-10}
      &\multirow{8}{*}{$<2\times10^3K$} &\textbf{A$^2$F-M(ours)}  & \textbf{1010k} & \textbf{56.7G}   & \textbf{32.28/0.8955}          & \textbf{28.62/0.7828}          & \textbf{27.58/0.7364}          & \textbf{26.17/0.7892}        & \textbf{30.57/0.9100}          \\
      && AWSRN-M      & 1254K & 72G     & 32.21/0.8954   & 28.65/0.7832   & 27.60/0.7368   & 26.15/0.7884 & 30.56/0.9093   \\
      && \textbf{A$^2$F-L(ours)}  & \textbf{1374K} & \textbf{77.2G}   & {\color{red} \textbf{32.32/0.8964} }         & {\color{blue}\textbf{28.67/0.7839}}          & {\color{blue}\textbf{27.62/0.7379}}          & {\color{red} \textbf{26.32/0.7931} }       & {\color{red}\textbf{30.72/0.9115}}          \\
      && SelNet       & 1417K & 83.1G   & 32.00/0.8931   & 28.49/0.7783   & 27.44/0.7325   & -            & -              \\
      && SRMDNF       & 1555K & 89.3G   & 31.96/0.8930   & 28.35/0.7770   & 27.49/0.7340   & 25.68/0.7730 & -              \\
      && AWSRN        & 1587K & 91.1G   & {\color{blue} 32.27/0.8960}   & {\color{red} 28.69/0.7843}   & {\color{red} 27.64/0.7385}   & {\color{blue} 26.29/0.7930} & {\color{blue} 30.72/0.9109}   \\
      && CARN         & 1592K & 90.9G   & 32.13/0.8937   & 28.60/0.7806   & 27.58/0.7349   & 26.07/0.7837 & -              \\
      && DRCN         & 1774K & 17974G  & 31.53/0.8854   & 28.02/0.7670   & 27.23/0.7233   & 25.14/0.7510 & 28.98/0.8816   \\ \cline{2-10}
      &\multirow{2}{*}{$<10^4K$}& SRDenseNet   & 2015K & 389.9K  & 32.02/0.8934   & 28.35/0.7770   & 27.53/0.7337   & 26.05/0.7819 & -              \\
      && MSRN         & 6078K & 349.8G  & 32.26/0.8960   & 28.63/0.7836   & 27.61/0.7380   & 26.22/0.7911 & 30.57/0.9103   \\
      \hline
  \end{tabular}}
  
  \label{quantityresults}
\end{table*}


\clearpage
\bibliographystyle{splncs}
\bibliography{0825}

\end{document}